\pdfoutput=1

\documentclass[11pt]{article}
\usepackage[table]{xcolor}  %
\usepackage{tabularray}
\UseTblrLibrary{booktabs} %
\definecolor{lightgray}{gray}{0.9}
\definecolor{lightblue}{RGB}{230, 240, 250} %
\definecolor{lightred}{RGB}{250, 230, 240} %

\usepackage[]{acl}
\usepackage{todonotes}
\usepackage{tablefootnote}
\usepackage{makecell}
\usepackage{multirow}
\usepackage{colortbl}
\usepackage{times}
\usepackage{latexsym}
\usepackage{tabularx}
\usepackage{booktabs}
\usepackage{array}
\usepackage{amsmath} %
\usepackage{amssymb} %
\usepackage{tcolorbox}
\usepackage{siunitx}
\usepackage{graphicx}
\usepackage{subfig}

\usepackage{listings}
\usepackage{listingsutf8}
\usepackage[T1]{fontenc}

\usepackage{microtype}

\title{Quantifying the Persona Effect in LLM Simulations}

\author{Tiancheng Hu\\
  University of Cambridge\\
  \texttt{th656@cam.ac.uk} \\\And
  Nigel Collier \\
  University of Cambridge\\
  \texttt{nhc30@cam.ac.uk} \\}

\begin{document}
\maketitle
\begin{abstract}

Large language models (LLMs) have shown remarkable promise in simulating human language and behavior. This study investigates how integrating persona variables—demographic, social, and behavioral factors—impacts LLMs' ability to simulate diverse perspectives. We find that persona variables account for <10\% variance in annotations in existing subjective NLP datasets. Nonetheless, incorporating persona variables via prompting in LLMs provides modest but statistically significant improvements. Persona prompting is most effective in samples where many annotators disagree, but their disagreements are relatively minor. Notably, we find a linear relationship in our setting: the stronger the correlation between persona variables and human annotations, the more accurate the LLM predictions are using persona prompting. In a zero-shot setting, a powerful 70b model with persona prompting captures 81\% of the annotation variance achievable by linear regression trained on ground truth annotations. However, for most subjective NLP datasets, where persona variables have limited explanatory power, the benefits of persona prompting are limited.\footnote{Code and data will be released at \url{https://github.com/cambridgeltl/persona_effect}} 
\end{abstract}

\section{Introduction}
Annotation questions such as ``how do you feel emotionally after reading this text'' are subjective - there are rarely definitive right or wrong answers~\cite{ovesdotter-alm-2011-subjective}. This subjectivity is increasingly being recognized within the NLP community. Subjective NLP tasks are typically characterized by low inter-annotator agreement, making label aggregation inappropriate~\cite{ovesdotter-alm-2011-subjective, plank-2022-problem,cabitza_toward_2023}. Previous research has established the significant influence of sociodemographic variables on the annotations of these tasks~\cite[\textit{inter alia}]{sap-etal-2022-annotators, santy-etal-2023-nlpositionality,pei-jurgens-2023-annotator}.

One approach to model these persona variables\footnote{In our work, we adopt a broad definition of \textit{persona variables} to include not only demographic and social variables but also other variables that could help describe a persona, such as variables relating to attitudes, behaviors, lived experiences, and values. It is worth noting that most NLP datasets have no information of any kind available about the annotators.
} is to use LLMs. LLMs have been effectively utilized for role-playing and simulating human behavior, primarily by defining the persona of interest within the prompt~\cite{aher_using_2023,horton2023large, kovac_large_2023, argyle_out_2023}. Their success has even spurred debates on whether LLMs could replace human subjects~\cite{Dillion2023,doi:10.1126/science.adi1778}.
However, there are also concerns about such ``persona prompting'' methodology (Figure \ref{fig:persona_prompting})~\cite{beck_how_2023}, citing ecological fallacy~\cite{orlikowski-etal-2023-ecological}, and LLMs' susceptibility to caricatures~\cite{cheng_compost_2023}, misportrayal and erasure of subgroup heterogeneity~\cite{DBLP:journals/corr/abs-2402-01908}.

Existing studies have often sought to measure the effects of individual persona variables, overlooking a holistic analysis of the potential explanatory power of persona variables on annotation variance. It is then hard to contextualize the models' ability to utilize persona information. To address this issue, our research explores the following questions:

\begin{figure*}[h]
\centering
\includegraphics[width=0.9\linewidth]{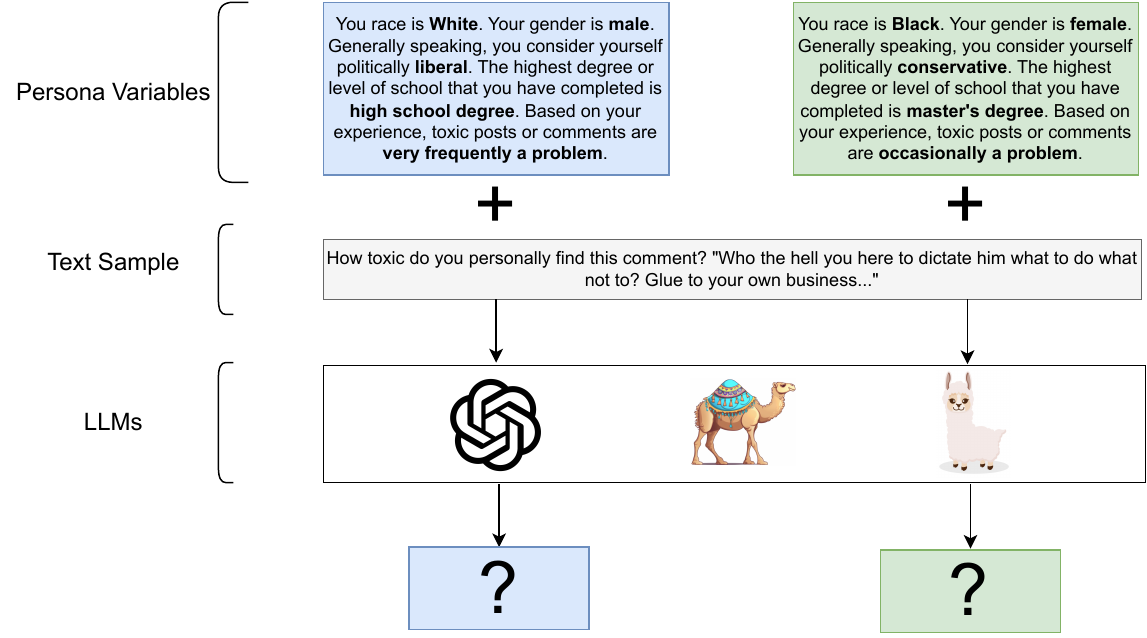}
\caption{Illustration of persona prompting. We prepend the persona information of an annotator before the text sample and task description to investigate the capacity of LLMs to simulate diverse perspectives in subjective NLP tasks.}
\label{fig:persona_prompting}
\end{figure*}
\noindent{\textbf{RQ1}: How much variance in human annotation could persona variables explain?} 
Understanding this will help us assess the overall influence of persona variables on human annotation, providing context to our subsequent investigations. We propose employing a linear regression analysis to predict annotations using persona variables and examine the resulting $R^2$ values. We find that persona variables explain relatively little variance (<10\%) for many NLP tasks (Section \ref{section:RQ1}). This general framework can be useful in understanding the potential effectiveness of LLM simulations prior to conducting large-scale experiments when some amounts of human data are available. 

\noindent{\textbf{RQ2}: Can incorporating persona variables via prompting improve LLMs' predictions?} Building on RQ1, we assess how much the explained variance by persona variables translates into prediction gains in LLMs. We find that incorporating persona variables provides modest but statistically significant improvements (Section \ref{sec:result_basic_setting}).

\noindent{\textbf{RQ3}: For what types of samples is persona prompting most useful?} To better understand the utility of persona prompting, we examine its impact across sample types, in terms of annotation entropy and standard deviation. We identify that most gains occur in samples characterized by frequent annotator disagreements within a relatively narrow range (high entropy-low standard deviation), suggesting that models can adjust their annotation to suit the persona, though not drastically (Section \ref{sec:case_study_1}).

\noindent{\textbf{RQ4}: How effectively can LLMs simulate personas when the importance of persona variables varies?}
Using a set of survey questions, where persona variables explain the responses to varying degrees, we apply persona prompting to LLMs. We find a linear relationship in our setting: the more persona variables are correlated with the outcome variable, the better LLMs predictions are using persona prompting. Large, preference-tuned models perform best and can explain up to 81\% of variance found in human responses. However, when the utility of persona variables is low, persona prompting has little effect. Regrettably, most subjective NLP datasets fall into this category, casting doubt on the efficacy of persona prompting in the current NLP context (Section \ref{sec:case_study_2}). Similar methodologies could be applied across different domains to better understand the simulation capabilities of LLMs.
\section{Related Work}
\label{sec:related_work}

\subsection{The Relationship between Persona Variables and Annotation Outcome}
The role of persona variables, such as demographics and lived experiences, in influencing annotations in NLP tasks is well established. Many studies have highlighted how persona variables affect tasks like hate speech detection~\cite{kumar_designing_2021, sap-etal-2022-annotators,pei-jurgens-2023-annotator,santy-etal-2023-nlpositionality,hettiachchi_how_2023,lee2023crehate}, sentiment analysis~\cite{ding_impact_2022, biester_analyzing_nodate}, and irony detection~\cite{frenda-etal-2023-epic}. While these studies shed light on the subjectivity of NLP annotations in many tasks, they often stop short of a holistic account of the  explanatory power of persona variables on annotation variance. By contrast, in social science, the impact of persona variables on attitude are long studied and quantified~\cite{bobo1989education, Bartels_2002}. In our work, we analyze the utility of the persona variables in explaining annotation outcomes across subjective NLP tasks. 

\subsection{Modeling Persona Variables and LLM for Simulation}
Some studies make use of persona variables only to enhance the diversity of model output, often without a strict emphasis on the accuracy or fidelity of persona representation in the output \cite{10.1145/3544548.3580688,park_generative_2023, liu2024toad}. Meanwhile, several other works emphasize the accuracy of persona representation and have sought to account for the differences between individual annotators or the group-level attributes of annotators through adding individual (group) specific layers~\cite[\textit{inter alia}]{davani-etal-2022-dealing,gordon_jury_2022,fleisig_when_2023,orlikowski-etal-2023-ecological}, or via prompting~\cite{beck_how_2023}. Results from these studies have been mixed, with some work indicating success using group-level persona variables~\cite{gordon_jury_2022,fleisig_when_2023}, while others cast doubt on the effectiveness of such methods \cite{orlikowski-etal-2023-ecological, cheng_compost_2023, beck_how_2023}. Simultaneously, in the social sciences, a multitude of studies have been employing use persona prompts in LLMs to simulate human behavior~\cite{horton2023large, argyle_out_2023,kim_ai-augmented_2023,tornberg2023simulating}, while others have pointed out the lack of fidelity and diversity in such simulations~\cite{bisbee_synthetic_2023, park2024diminished,DBLP:journals/corr/abs-2402-01908,taubenfeld2024systematic}.

Our work builds on the uncertainty raised by these mixed results, focusing on the potential of persona prompting with LLMs for simulating different perspectives in NLP tasks, which is currently understudied. Furthermore, our work aims to isolate the evaluation of \textit{persona} prompting from the impact of \textit{text samples} in the modeling process, a separation that has not been much explored in previous studies.

\subsection{Persona Prompting and AI Alignment}
\label{sec:alignment}
Apart from the research focused on incorporating demographic factors into NLP models and using LLMs for simulations, another line of studies has examined persona prompting in the context of AI alignment~\cite{santurkar_whose_2023,durmus_towards_2023}. These studies have employed LLMs to answer multiple-choice survey questions concerning societal values and attitudes, comparing the LLM-generated answer distribution with actual human response distribution derived from survey data representing diverse demographic groups. In contrast to these studies, our work aims to explore the efficacy of LLMs in leveraging persona variables to inform task predictions, rather than the degree to which LLM responses to survey questions mirror those of specific demographic groups.

\section{RQ1: How much variance in human annotation could persona variables explain?}
\label{section:RQ1}

\paragraph{Methodology} Given the relative gap in literature in a holistic understanding of the impact of persona variables on annotation variance, we investigate to what extent persona variables explain human annotation variance. This analysis would provide valuable context to any modeling exercise of incorporating persona variables.

We employ a mixed-effect linear regression model\footnote{In R notation, \texttt{annotation $\sim$ persona variables + (1 | text\_id)}} to assess how much variance in annotation can be explained by persona variables (fixed effect), while controlling for the text-specific variability in the text sample (random effect) by fitting a random intercept for each text. Using a mixed-effect linear regression allows us to separate the impact of persona variables from the inherent variation of the text being annotated. We also consider incorporating an additional random effect term to capture individual annotator differences; however, the fixed effect estimates are very similar. Consequently, we opt to include only a random effect for the text sample. We evaluate 10 subjective NLP datasets which provide unaggregated annotations and annotator persona variables. We also consider the presidential vote question in the ANES 2012 public opinion survey~\cite{anes2012}, in which every human subject answers the same question and therefore does not require a text random effect, for comparison.

\paragraph{Results} We show
a comparison of the tasks, sources of data, annotation methods, sizes, types of persona information included, and the regression $R^2$ values in Table \ref{tab:rq1_r2}. 
\begin{table*}[!h]
\resizebox{\textwidth}{!}
{
    \begin{talltblr}[label=none, note{a} = {Another phase of this dataset has 600+ annotators labeling a total of 15 tweets.},note{b}={We consider the action acceptability, to be in line with the NLPositionality dataset. As it is a volunteer-annotated dataset, substantial persona information is unavailable.},note{c}=After filtering out participants with missing attributes.]{
      colspec={lccccccc},
      row{odd}={bg=},
      row{even}={bg=lightgray}
    }
    \toprule
    \textbf{Task} & \textbf{Dataset} & \textbf{Data Source} & \textbf{Annotation} & \textbf{Size} & \textbf{Persona\\Variables} & \textbf{$R^2_{Cond.}$} & \textbf{$R^2_{Marg.}$}\\
    \midrule
    Toxicity Detection & {annWithAttitudes\\\cite{sap-etal-2022-annotators}} & Twitter &  {5-point\\MTurk} &  {$N$=626\\$A$=5.5\TblrNote{a}\\U.S.}&  {Basic\\Attitude} & 0.611 & 0.045 \\
    Offensiveness Rating &  {POPQUORN\\\cite{pei-jurgens-2023-annotator}} & Reddit &  {5-point\\Prolific}& {$N$=1,500\\$A$=8.7\\U.S.}&Basic&0.319&0.029 \\
    Politeness Rating  &  {POPQUORN\\\cite{pei-jurgens-2023-annotator}} & Email & {5-point\\Prolific} & {$N$=3,718\\$A$=6.7\\U.S.}& Basic  &  0.454 & 0.014\\
    Toxicity Detection &\citet{kumar_designing_2021} &  {Twitter\\Reddit\\4chan} & {5-point\\MTurk} & {$N$=106,035\\$A$=5.1\\U.S.}&  {Basic\\Attitude\\Behavior} & 0.349 & 0.106\\
    Sentiment Analysis &\citet{diaz_addressing_2018}  & Twitter & 5-point & {$N$=14,071\\$A$=4.2\\U.S.}&  {Basic\\Attitude}  & 0.329 & 0.036\\
    Social Acceptability& {Social-Chem-101\\~\cite{forbes-etal-2020-social}} & Reddit & {5-point\\MTurk}& {$N$=9,740\\$A$=6.1\\Mostly U.S.}& Basic & 0.432 & 0.097\\
    Social Acceptability &  {NLPositionality\TblrNote{b}\\\cite{santy-etal-2023-nlpositionality}} & Reddit &  {5-point\\Opt-in volunteer} & {$N$=291\\$A$=50.2\\87 countries}& Basic & 0.513 & 0.005\\
    Toxicity Detection &  {NLPositionality\TblrNote{b}\\\cite{santy-etal-2023-nlpositionality}} & Twitter &   {3-point\\Opt-in volunteer} & {$N$=299\\$A$=29.6\\87 countries}& Basic & 0.432 & 0.017\\
    Social Bias& {SBIC\\\cite{sap-etal-2020-social}}&  {Twitter\\Reddit\\Gab\\Stormfront} &  {3-point\\MTurk} & {$N$=35,504\\$A$=3.2\\U.S. and Canada} & Basic & 0.758 & 0.031 \\
    Irony Detection& {EPIC\\\cite{frenda-etal-2023-epic}} &  {Twitter\\Reddit} &  {Binary\\Prolific} & {$N$=2,994\\$A$=4.7\\IE, UK, US, IN, AU}& Basic & 0.289 & 0.091 \\
    Presidential Vote& \citealp{anes2012} 2012& Survey & {Binary\\Face-to-face}  &  {$A$=2,728\TblrNote{c}\\U.S.} &  {Basic\\Attitude\\Behavior} & - & 0.719 \\
    
    \bottomrule
    \end{talltblr}
}

\caption{An overview of datasets with unaggregated annotations and persona information. This table compares the tasks, sources of data, annotation methods, sizes, types of persona information included, and to what degree the persona variables can explain the variance of annotations in each dataset. The ``Size'' column specifies the number of text samples ($N$) and the average number of annotators per sample ($A$), alongside the geographical location of the annotators. The ``Persona Variables'' column indicates the available persona categories: ``Basic'' for standard demographics like gender and age, ``Attitude'' for annotators' personal views, and ``Behavior'' for actions such as media consumption habits. The conditional ($R^{2}_{Cond.}$) and marginal ($R^{2}_{Marg.}$) R-squared values are reported from regression models that predict the annotations based on persona variables, while accounting for text-specific variability (using a random effect for each text).}
\label{tab:rq1_r2}
\end{table*}

We observe that the datasets mostly come from social media sources and annotations are collected through crowd-sourcing. They vary substantially in size, persona variables provided and $R^2$ values. While persona variables (fixed effect) do significantly explain some variance in annotation outcomes, they account for just \textbf{1.4\%-10.6\%} of the total variance (Marginal $R^2$), even when controlling for text variation. Conversely, variability inherent to individual texts (random effect) can explain up to \textbf{70\%} of the total variance, i.e. $\sim(\text{Conditional } R^2 - \text{Marginal } R^2)$. For comparison, in the ANES dataset, persona variables explain more than \textbf{70\%} human response variance.

The marginal $R^2$ values provide a baseline indication of the variance in annotations that persona variables could explain. The regression model assumes a linear relationship between persona variables and annotation and does not consider any interaction between the persona variables. Therefore, while it is straightforward and interpretable, for LLMs, it should be considered a \textbf{weak baseline}. 

Acknowledging that a substantial portion of variance remains unexplained (\textbf{25\%-70\%}) by either the text or persona variables across all tasks considered is crucial. This unexplained variance could be attributed to theoretically measurable persona factors such as personality traits and complex moral and political beliefs, which are not currently collected in existing datasets. Additionally, it could be due to hard-to-measure factors like the annotators' lived experiences, interpersonal dynamics, and other personal variables.

The elevated $R^2$ value in the ANES dataset may be attributed to the escalating degree of polarization in U.S. politics in recent years. This rise in polarization has lead to more predictable voting patterns~\cite{pew2014political} and the increasing tendency of U.S. voters to behave in a manner consistent with their in-groups~\cite{graham2010beyond}. 

In contrast, tasks such as assessing the hatefulness of a tweet offer more room for personal interpretation, leading to diverse opinions. Thus, persona factors may account for a lesser portion of the variance in annotation for such tasks.

We argue that regression analysis offers a valuable framework for setting realistic expectations regarding the fidelity of persona prompting with LLMs. Specifically, when some level of annotated data is available, this approach offers preliminary insights into potential simulation results, allowing researchers to gauge the likely performance of persona-prompted LLMs for a new application. This then enables an informed decision-making process before committing to costly large-scale simulation runs.

\section{RQ2: Can incorporating persona variables via prompting improve LLMs' predictions?}
\label{sec:result_basic_setting}
\paragraph{Methodology}
Since persona variables can explain a small but significant amount of human annotation variations, we then explore whether persona prompting would improve LLM's predictions. 

As depicted in Figure \ref{fig:persona_prompting}, we prepend each \textit{text sample} with \textit{persona variables} in a zero-shot prompting setup. We prompt the LLMs twice: once with persona variables, and once without, to zero-shot predict individual annotations on AnnotatorwithAttitude~\cite{sap-etal-2022-annotators},~\citet{kumar_designing_2021}, EPIC~\cite{frenda-etal-2023-epic} and the politeness rating task in POPQUORN~\cite{pei-jurgens-2023-annotator}. We preserve the original language of the persona descriptions to the extent possible, adopt a multiple-choice format, include a description of the question and the answer choices, and predict only the next token as the model's response, as done in prior work~\cite{santurkar_whose_2023,durmus_towards_2023}. Due to cost constraints, we sample 600 instances from each dataset. The details of the prompt format are provided in the Section \ref{sec:prompt_template}.

We additionally perform a set of robustness experiments by swapping the order of persona variables in the prompt or paragraphing the language used to describe each persona variables and repeat the experiments on~\citet{kumar_designing_2021}. The detailed setting can be found in Section \ref{sec:robustness}.

To evaluate, we compare model predictions with individual human annotations using $R^2$ value, Cohen's Kappa~\cite{cohenskappa}, mean absolute error (MAE) for multi-class classification or macro F1 score for binary classification, given the class imbalance in~\citet{frenda-etal-2023-epic}. Our primary objective is to observe the performance changes induced by persona prompting, rather than focusing on the absolute performance of each model.
\begin{table*}[h]
    \centering
    \resizebox{\textwidth}{!}{%
        \colorbox{white}{%
    \begin{tabular}{>{\kern-\tabcolsep}lcccccccccccc<{\kern-\tabcolsep}}
        \toprule
        \textbf{Model} & \multicolumn{3}{c}{\textbf{annwAttitudes}} & \multicolumn{3}{c}{\textbf{\citet{kumar_designing_2021}}} & \multicolumn{3}{c}{\textbf{EPIC}} & \multicolumn{3}{c}{\textbf{POPQUORN-P}} \\
        
        \cmidrule(lr){2-4} \cmidrule(lr){5-7} \cmidrule(lr){8-10} \cmidrule(lr){11-13}
        & \boldmath{$R^2\uparrow$} & \boldmath{$\kappa\uparrow$} & \textbf{MAE ↓} & \boldmath{$R^2\uparrow$} & \boldmath{$\kappa\uparrow$} & \textbf{MAE ↓} & \boldmath{$R^2\uparrow$} & \boldmath{$\kappa\uparrow$} & \textbf{F1 ↑} & \boldmath{$R^2\uparrow$} & \boldmath{$\kappa\uparrow$} & \textbf{MAE ↓} \\
        
        \midrule
        Target & 0.64 (0.03) & - & - & 0.42 (0.20) & - & - & 0.28 (0.09) & - & - & 0.47 (0.03) & - & - \\ \midrule
        GPT-4-0613& 0.56 & 0.42 & 0.70 &0.16&0.24&0.87&0.03 & 0.12 & 0.52 & 0.34 & 0.22 & 0.89\\
       \rowcolor{lightgray}\quad+Persona& 0.53 & 0.40 & 0.74 &0.12 & 0.20 & 0.90 & 0.05*& 0.20*& 0.58* & 0.33 & 0.22 & 0.90 \\
        GPT-3.5-Turbo-0613 & 0.53 & 0.29 & 0.80 & 0.12 & 0.17 & 1.12&0.04&0.18&0.59 & 0.28 & 0.09 & 1.07 \\
        \rowcolor{lightgray}\quad+Persona & 0.49 & 0.31 &0.82 & 0.12 & 0.15 & 0.97*&0.03&0.14&0.54 & 0.28 & 0.14* & 1.14\\
        Llama-2-70b & 0.17 & 0.14 & 1.70 & 0.01 & 0.04 & 1.51 & 0.00 & 0.00 & 0.24 & 0.24 & 0.13 & 1.42 \\
       \rowcolor{lightgray}\quad+Persona & 0.40* & 0.30* & 0.91* & 0.03 & 0.05 & 1.01* & 0.00 & 0.00 & 0.24 & 0.21 & 0.17 & 1.10* \\
        Llama-2-70b-chat & 0.39 & 0.13 & 1.33 & 0.11 & 0.07 & 1.70 & 0.00 & 0.05 & 0.49 & 0.32 & 0.15 & 1.00 \\
       \rowcolor{lightgray}\quad+Persona & 0.42 & 0.15 & 1.22* & 0.10 & -0.01 & 1.45* & 0.02* & 0.14* & 0.56* & 0.31 & 0.14 & 0.90* \\
        Tulu-2-70b & 0.49 & 0.29 & 0.90 & 0.16 & 0.13 & 1.09 & 0.05 & 0.20 & 0.59 & 0.34 & 0.20 & 0.89 \\
        \rowcolor{lightgray}\quad+Persona & 0.49 & 0.26 & 0.88 & 0.14 & 0.16 & 0.90* & 0.07 & 0.27* & 0.63* & 0.31 & 0.16 & 0.92 \\
        Tulu-2-dpo-70b & 0.51 & 0.35 & 0.84 & 0.15 & 0.15 & 1.16 & 0.03 & 0.14 & 0.54 & 0.35 & 0.21 & 0.83 \\
        \rowcolor{lightgray}\quad+Persona & 0.51 & 0.30 & 0.84 & 0.15 & 0.20* & 0.92* & 0.04 & 0.18 & 0.58* & 0.33 & 0.19 & 0.87 \\
        \midrule
        Avg. \(\Delta \) & 0.06* & 0.02 & -0.14* & -0.01 & -0.01 & -0.22* & 0.01* & 0.04* & 0.02* & -0.02 & 0.00 & -0.05*\\
        \bottomrule

    \end{tabular}
    }}
    \caption{Comparison of performance across LLMs in estimating individual annotations, with and without the inclusion of persona variables. Performance is measured using $R^2$, Cohen's Kappa ($\kappa$), Mean Absolute Error (MAE) and macro F1 score. Asterisks (*) denote statistically significant improvements when persona variables are included.}
    \label{tab:RQ1_result}
\end{table*}
\paragraph{Result}
We show the results in Table \ref{tab:RQ1_result}. The first row shows the ``Target'' $R^2$ values, which refer to the conditional (and marginal) $R^2$ value of the mixed-effect regression on the sampled data computed as in Table \ref{tab:rq1_r2}, while the $R^2$ in subsequent rows are from a fixed-effect linear regression predicting the human annotation with model predictions\footnote{in R notation, \texttt{annotation $\sim$ prediction}}. While these two $R^2$ values cannot be compared directly, the ``Target'' $R^2$ gives context to the fixed-effect $R^2$ values. To evaluate the statistical significance of performance differences between models incorporating and excluding persona variables, we conduct a bootstrap analysis~\cite{Efron1992} with 1,000 replications. We denote with asterisks those instances where incorporating persona variables leads to statistically significant performance improvements. For the overall improvement (last row), we aggregate predictions from the 6 models and apply the same bootstrapping procedure to assess the collective effect of persona variables. As the 7b and 13b models exhibit much weaker performance, we only feature results from 70b models in the main text, while the results from smaller models are included in Table \ref{tab:RQ1_result_full}.

At the aggregate level, persona prompting provides varying levels of statistically significant improvement across at least one metric in each of the four datasets. However, these improvements are generally modest. For instance, in EPIC, where persona variables could explain up to 9\% of annotation variance, persona prompting only provides 1\% gain on average. The effectiveness of persona prompting also varies across models: for each dataset, persona prompting improves the performance of some models but not others, echoing the results in~\citet{beck_how_2023}. 

We note that overall, with and without persona prompting, GPT-4 consistently outperforms all other models in every task. Tulu-2 models outperform Llama-2 with performance on par with GPT-3.5. The Llama-2 models are, on the other hand, much more sensitive to persona variables, arguably to an excessive degree. For example, on AnnotatorwithAttitudes, persona prompting improves the $R^2$ by as much as 0.23 even though persona variables only has a marginal Target $R^2$ of 0.03.
We show the robustness experiment result in Table \ref{tab:robustness}. The model performances are consistent across variations in the ordering and language use of the persona variables.

\section{RQ3: For what types of samples is persona prompting most useful?}
\label{sec:case_study_1}
\paragraph{Methodology}
To better understand persona prompting as a technique, we aim to investigate its effectiveness on data samples with varying degrees of annotation \textit{entropy} and \textit{standard deviation}. We focus on~\citet{kumar_designing_2021}, as persona variables play a relatively more important role in explaining annotation variances in this dataset.

We create a new subsample of the dataset with four categories: low entropy-low standard deviation (most annotators agree with one another and the magnitude of the disagreement is small, e.g. 1, 1, 1, 1, 2); low entropy-high standard deviation (e.g. 0, 4, 4, 4, 0); high entropy-low standard deviation (e.g. 1, 1, 2, 2, 3); and high entropy-high standard deviation (e.g. 0, 1, 2, 3, 4). The low/high division is based on the medians of entropy and standard deviation.
Then, we further stratify samples from each category into four bins according to their average annotation value. We then randomly sample 150 from each bin, culminating in a total of 600 samples per category. This approach is implemented to mitigate extreme class imbalances within certain categories. For instance, the low entropy-low standard deviation category would predominantly include samples with a rating of 0 (Not at all toxic). We then run the LLMs twice, once with persona prompting, once without, in the same setting as described in Section \ref{sec:result_basic_setting}, on Llama-2-70b, Llama-2-70b-chat, Tulu-2-70b, and Tulu-2-dpo-70b.

\begin{figure*}[ht]
    \centering
    \subfloat[\citet{kumar_designing_2021}]{
        \includegraphics[width=0.48\linewidth]{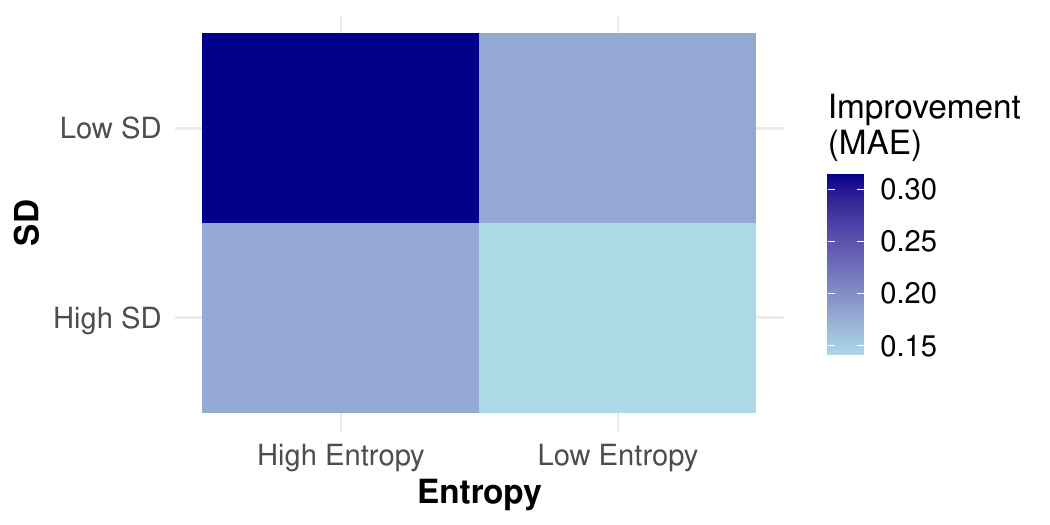}
        \label{fig:case_study_1_sub1}
    }\hfill
    \subfloat[POPQUORN-P]{
        \includegraphics[width=0.48\linewidth]{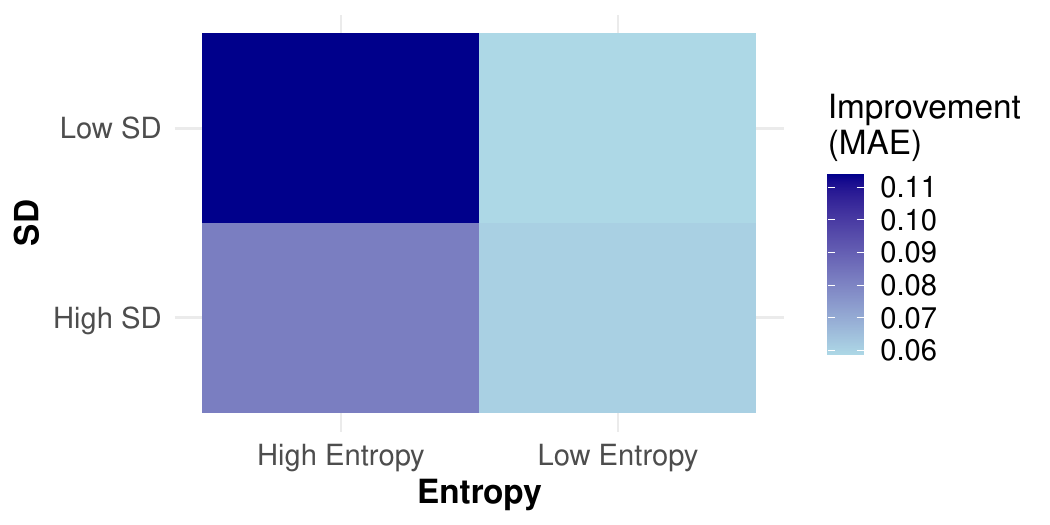} %
        \label{fig:case_study_1_sub2}
    }
    \caption{Mean improvement in MAE with persona prompting across four 70b models in annotations characterized by low/high entropy and standard deviation, with darker colors denoting more substantial improvement in predictions.}
    \label{fig:case_study_1}
\end{figure*}

\paragraph{Result} We show in Figure \ref{fig:case_study_1_sub1} the mean improvement in MAE between models with and without persona prompting, averaged across the four models, in each of the four categories, with darker color indicating a greater degree of improvement in predictions when persona prompting is used. To reduce the possibility of finding a dataset-specific effect, we also repeat the same experiment on POPQUORN-Politeness dataset \cite{pei-jurgens-2023-annotator}, and show the same plot Figure \ref{fig:case_study_1_sub2}.

Our findings indicate that including persona information leads to only slight changes in the model's predictions for data with low entropy. This is as expected - with or without persona prompting, a capable LLM should already capture the consensus among annotators if there is one, thus only necessitating minor adjustments to individual predictions.

On the contrary, we observe larger shifts in prediction when annotations have high entropy but low standard deviation. These instances often involve substantial disagreement among individuals, though within a small margin. The integration of persona variables may then enhance the model's ability to refine its predictions. An example in this would be a prediction transition from 3 (without persona variables) to 4 (with persona variables).

However, when both entropy and standard deviation are high, the task of adjusting predictions based on persona information becomes considerably more challenging, as this would require significant shifts in the predicted values from the ``mean'' level, when no persona variables are provided. For instance, imagine a case where a prediction needs to change from 0 (without persona variables) to 4 (with persona variables).

While varying in magnitude, the MAE improvements when including persona variables are significant in all four categories in both datasets, based on bootstrapping 1,000 runs. Additionally, to determine whether there are statistically significant differences in improvements across the four categories, we perform a one-way ANOVA and find significant differences in the improvements. Finally, we conduct Tukey's range test for each pair of categories. For \citet{kumar_designing_2021}, the high entropy-low standard deviation category statistically significantly outperforms the other three categories, while the differences between the other three categories are not significant. For the POPQUORN-Politeness dataset, the high entropy-low standard deviation setting consistently shows the most improvement. While not all comparisons reach statistical significance, it never performs significantly worse than any other category. High entropy-high standard deviation and low entropy-low standard deviation show some variation in performance in terms of statistical significance, but neither shows more improvements than high entropy-low standard deviation setting. Low entropy-high standard deviation consistently yields statistically the least MAE improvements in all comparisons.

\section{RQ4: How effectively can LLMs simulate personas when the importance of persona variables varies?}
\label{sec:case_study_2}
\paragraph{Motivation} Within the context of NLP annotation, both the \textit{text sample} and the \textit{persona variables} may vary across instances (Figure \ref{fig:persona_prompting}). Both factors,   along with their interactions, could potentially influence model predictions. To understand the models' capacity for simulating different perspectives with persona prompting, we designed a case study that minimizes the impact of the \textit{text sample}.

\paragraph{Methodology}
We use the ANES dataset~\cite{anes2012}, a comprehensive U.S. national-level election survey, as a data source for this section. This dataset offers a wealth of persona variables from a large sample of respondents. From the perspective of NLP annotation, surveys can be seen as having a large number of individuals (typically >1,000) annotating a small number of sentences, each representing a question. One key difference is that the survey questions, carefully crafted and tested by seasoned professionals, are designed to eliminate ambiguity common in social media-based NLP text annotation datasets. Therefore, by running experiments on the ANES dataset, we can minimize the impact of the randomness in the text samples.

We select a number of questions from ANES 2012 as the \textit{text sample}, or the questions to be predicted, using a fixed set of persona variables. We ensure that these questions have varying predictability from persona variables, indicated by $R^2$ values. Further details of the dependent and independent variables considered are included in Section \ref{sec:details_on_case_study_2}. After filtering out respondents who did not answer some of the questions of interest and performing random downsampling, we arrive at a sample size of 600 human respondents and 21 questions. Each question is paired with two levels of persona variables, resulting in 42 combinations of persona variables and questions, each with a different level of target $R^2$ (see Section~\ref{sec:details_on_case_study_2} for details). We then run the LLMs with persona prompting.

We also perform a robustness check with the presidential vote prediction question from ANES by swapping the order of persona variables in the prompt or paragraphing the language used to describe each persona variables. The detailed setting can be found in Section \ref{sec:robustness}.
\paragraph{Result}
\begin{figure*}[ht]
    \centering
    \subfloat[Tulu-2-dpo-70b]{
        \includegraphics[width=0.45\linewidth]{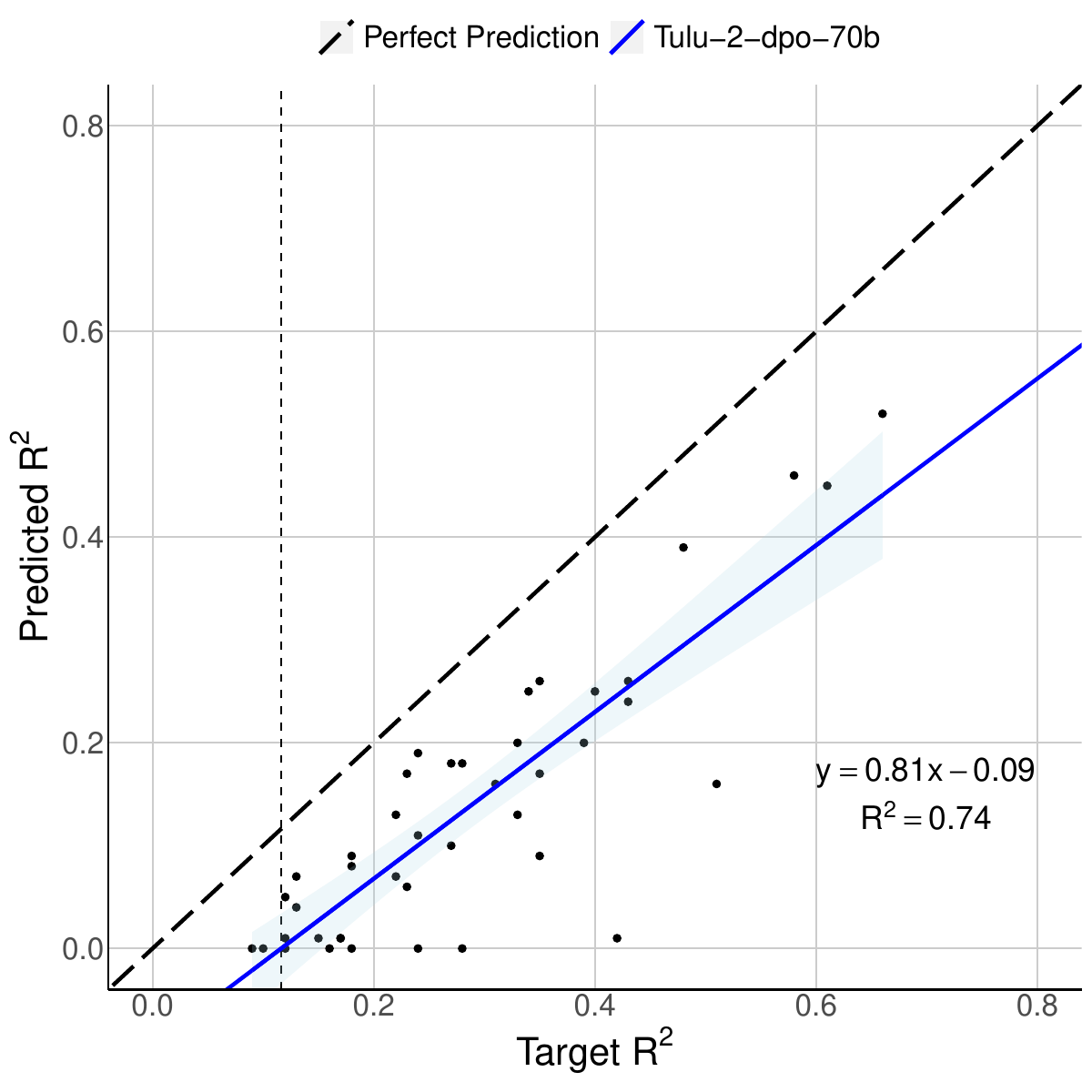}
        \label{fig:case_study_2_tulu2dpo70b}
    }
    \hfill
    \subfloat[Llama-2-7b-chat]{
        \includegraphics[width=0.45\linewidth]{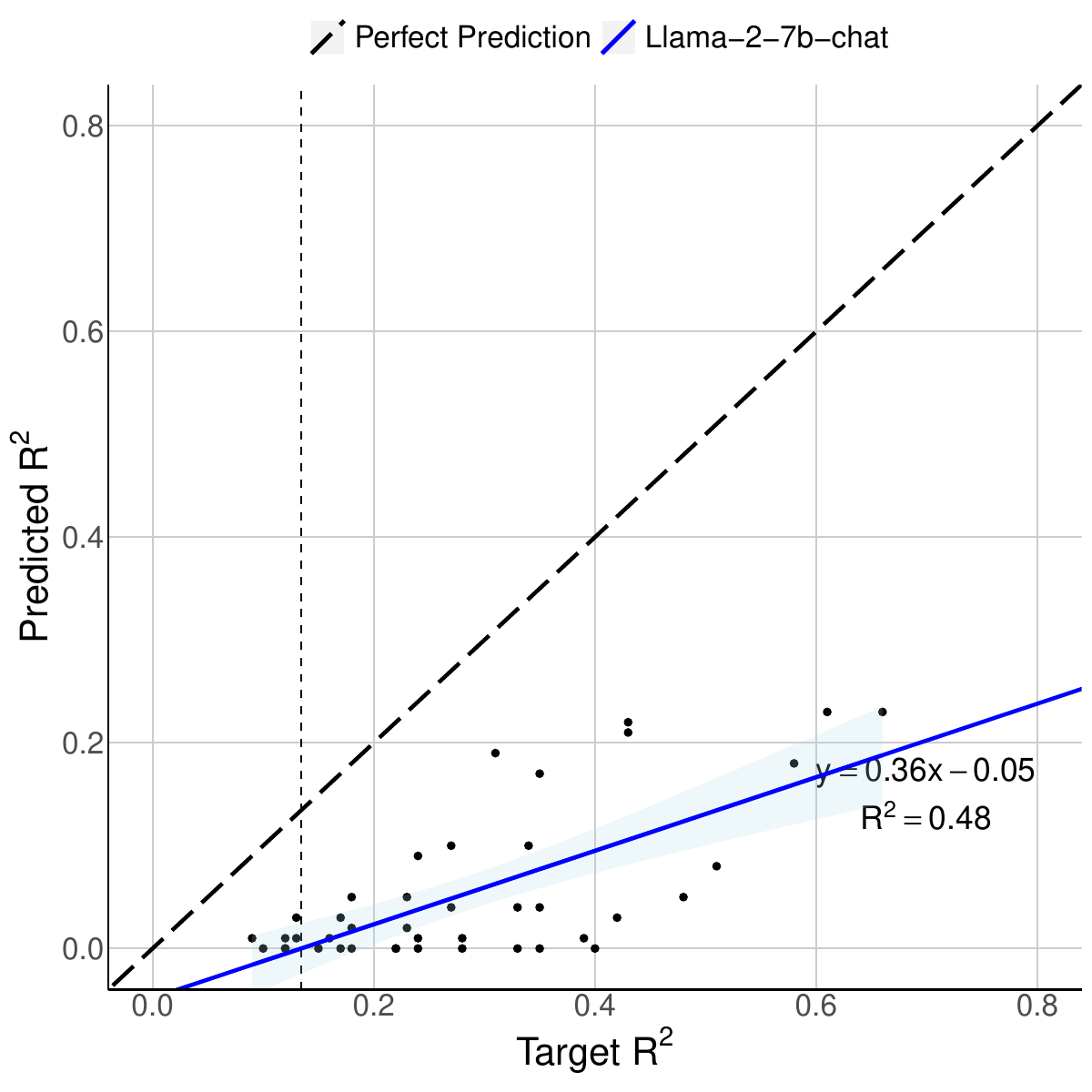} %
        \label{fig:case_study_2_tulu2dpo70b1}
    }
    \caption{Comparison of predicted $R^2$ and target $R^2$. Each point in the X-Y plane represents an experimental result with persona prompting, where the x-coordinate signifies the target $R^2$ and the y-coordinate denotes the predicted $R^2$. We then fit a linear regression line and also plot the maximum linear regression model performance line $y=x$ in the same figure.}
        \label{fig:case_study_2}
\end{figure*}

We visualize the relationship between the predicted and target $R^2$ values in Figure \ref{fig:case_study_2} of Tulu-2-70b-dpo and Llama-2-7b-chat. The results for other models are provided in the Figure \ref{fig:case_study_2_appendix}.  
Each point in the scatter plot represents an experiment result, where the x-coordinate signifies the target $R^2$ and the y-coordinate denotes the predicted $R^2$. The line $Y=X$ is also included to represent the maximum linear regression model performance, where predicted $R^2$ equals target $R^2$. We additionally fit a linear regression line to the data points and show the fitted equation and $R^2$ in the figure. 

Our results show a positive correlation between the target and predicted $R^2$ values - the higher the target $R^2$ value, the higher the predicted $R^2$. Tulu-2-70b-dpo, one of the best-performing models on the 70b scale, can capture 81\% of the target $R^2$. The other 70b models, except for the base model Llama-2-70b (Figure \ref{fig:case_study_2_appendix}), have similar simulation capabilities, while the smaller models (7b and 13b) do much worse. However, it is important to note that no model surpasses the $y=x$ line, suggesting that persona prompting still falls short compared to a trained linear model. The complex relationship between persona and target variables, including interaction terms, implies that the true target $R^2$ is likely much higher. Furthermore, even the best models fail to utilize the persona information effectively when target $R^2$ is low, especially when $R^2 < 0.1$. Nevertheless, when persona variables are sufficiently predictive of target variables, persona prompting can lead to somewhat accurate simulations on large models. 

Considering that most existing NLP datasets, as discussed in Section \ref{tab:rq1_r2}, have marginal $R^2 < 0.1$, we argue that \textbf{persona prompting cannot reliably simulate different perspectives within existing NLP tasks}. This finding may explain the modest gain of persona prompting observed earlier in Section \ref{tab:RQ1_result} and in~\citet{beck_how_2023}.

We propose two potential explanations as to why LLMs, however powerful they are in other tasks, may be deficient in simulating diverse perspectives: 

1) The persona variables typically accessible to researchers are group-level, while people form their identity based on both individual and group-level characteristics~\cite{10.1145/3290605.3300565}. Therefore, there could be an inherent mismatch between the group-level variables we provide and individual perspectives we aim to simulate.

2) LLM generations can be understood as simulating the medium of a group, rather than individuals~\cite{west2023generative}. Therefore, LLMs can have the tendency to represent a group as a monolith in simulation, thereby failing to capturing the inherent within-group heterogeneity~\cite{DBLP:journals/corr/abs-2402-01908}. While using more fine-grained group-level persona variables may in theory bring us closer to individual ratings, it remains to be seen whether this could lead to true individualization in practice.

We show the robustness experiment result in Table \ref{tab:robustness}. The model performances are consistent across variations in the ordering and language use of the persona variables descriptions.

\section{Conclusion and Recommendation}
\label{sec:discussion}
Our study reveals that persona variables account for less than 10\% of variance in human annotations across most NLP datasets we consider. The use of persona prompting offers modest yet significant improvements across different tasks. The improvement is most pronounced in cases where the annotators largely disagree but only by a small margin (high entropy-low standard deviation). By running a case study with U.S. opinion survey data, we uncover a linear relationship between target and predicted $R^2$ values. Alarmingly, when the target $R^2$ value falls below 0.1, the predicted $R^2$ often drops to zero. This could explain the small improvements observed in NLP tasks with persona prompting, as existing datasets often have $R^2$ values smaller than 0.1.

Based on these insights, we offer the following recommendations:

1) \textbf{Exercise Caution in LLM-Based Simulations:} In light of our findings, we advise caution for researchers intending to use LLMs for simulation purposes, especially in NLP tasks where persona variables' influence is likely weak (low target $R^2$). If the goal is to merely improve model generation diversity, without prioritizing the fidelity of the model output towards the persona variables, applying persona prompting as is may suffice. However, if the goal is to faithfully simulate human behavior, achieving high fidelity could be challenging. Unvalidated, zero-shot simulations with LLMs may not yield reliable results. Therefore, thorough validation and potentially fine-tuning are essential to ensure simulation fidelity. 

2) \textbf{Implement More Strategic Dataset Design:} The collection of persona information should be driven by clear objectives. If the aim is to understand how different groups annotate data, collecting only demographic information might be adequate but limited in scope for generalization of findings beyond the specific dataset. For behavioral simulation, a careful selection of persona variables is needed to increase the target $R^2$ and achieve better predictability. Future datasets could include more nuanced and targeted questions probing individual characteristics such as attitudes, beliefs, and behaviors. Moving forward, it is crucial to expand dataset collection efforts to encompass diverse cultural perspectives and multiple languages, especially those from non-U.S. contexts, to make language technologies more equitable globally.

\section{Limitations}
\label{sec:limitation}
We recognize the inherent subjectivity in human behavior and the multitude of contextual factors that influence decision-making, many of which are difficult to quantify. While incorporating more fine-grained persona variables could potentially reduce error margins significantly, some level of error is likely to persist. Furthermore, when collecting fine-grained persona information, researchers should carefully consider the ethical implications. In crowdsourcing environments, there is also a risk of obtaining intentionally inaccurate responses to sensitive questions~\cite{huang-etal-2023-incorporating}.

While we exerted considerable effort to include a diverse range of datasets, the vast majority of available datasets with persona information from annotators have been collected in the U.S., featuring persona questions primarily relevant to this particular context. Consequently, we can only speculate about the effectiveness of persona prompting for questions that are specifically tailored to other countries. Furthermore, to the best of our knowledge, as of the time of writing this paper, we have not identified any publicly available datasets that include annotator persona variables in a language other than English. Considering that even the most sophisticated LLMs still exhibit significant performance disparities between English and non-English languages~\cite{ahuja-etal-2023-mega}, it is highly probable that the ability of LLMs to simulate different perspectives based on persona information is considerably weaker in non-English languages. Additionally, many terms used to denote identities are deeply rooted in specific cultural and societal contexts, which cannot be readily translated into other languages. Thus, it is crucial to evaluate the simulation capabilities of an LLM independently for each language, without translation. 

The zero-shot simulation ability of LLMs largely depends on their extensive training data, essentially a compressed digital snapshot of the internet. However, previous studies have indicated that the pretraining corpora used by LLMs are riddled with various social biases~\cite[\textit{inter alia}]{thepile, dodge-etal-2021-documenting,doi:10.1126/sciadv.abm2463,hu2023generative}. Consequently, LLM simulations could potentially be tainted by biases and stereotypes, among other issues. 

We did not carry out extensive prompt engineering due to computational limitations and the targeted scope of our study. Instead, we presented the same prompts with persona information using language that closely mirrors how questions were asked of human participants. We believe this constitutes a fair setting for comparing LLMs. Additionally, we conducted a robustness check and found little variation for different persona variable orders and the exact wordings used to describe each variable (Section \ref{sec:robustness}).

\section{Ethical Considerations}
\label{sec:ethics}
We utilize persona variables from publicly available datasets, which have been anonymized prior to their release. Therefore, no human participants were involved or personal data collected in this study. The research acknowledges the potential risks associated with the use of LLMs for simulation purposes, including issues such as identity fraud and manipulation. We sternly denounce such nefarious applications of this technology. 
We also acknowledge the concerns related to categorizing individuals into different demographic groups. However, we argue that our study merely utilizes existing datasets and does not involve any original data collection. Furthermore, the categorizations employed within these datasets adhere to established best practices, such as those used by the U.S. Census Bureau, thereby ensuring their appropriateness. In addition, the use of these demographic categories is only aimed at understanding and demonstrating the potential for LLMs to simulate diverse perspectives.

\section*{Acknowledgements}
T.H is supported by Gates Cambridge Trust (grant OPP1144 from the Bill \& Melinda Gates Foundation). This work was performed using resources provided by the Cambridge Service for Data Driven Discovery (CSD3) operated by the University of Cambridge Research Computing Service (www.csd3.cam.ac.uk), provided by Dell EMC and Intel using Tier-2 funding from the Engineering and Physical Sciences Research Council (capital grant EP/T022159/1), and DiRAC funding from the Science and Technology Facilities Council (www.dirac.ac.uk).
We thank Chen Cecilia Liu, Songbo Hu, Alan Ansell, Philipp Borchert, Manoel Horta Ribeiro, Dirk Hovy, Zihao Fu, Yara Kyrychenko, Yijiang Dong, Shun Shao, Wen Wu, Meiru Zhang and Yinhong Liu for helpful feedback and discussions at various stages of the project.

\bibliography{anthology,custom}
\bibliographystyle{acl_natbib}

\appendix
\section{Implementation Detail}
In this work, we make use of the following pretrained LLMs: GPT-4~\cite{achiam2023gpt}, GPT-3.5~\cite{openai_introducing_nodate}, Llama-2~\cite{llama2}, Llama-2 Chat~\cite{llama2}, Tulu-2~\cite{ivison2023camels}, and Tulu-2-dpo~\cite{ivison2023camels}. We run all experiments on a NVIDIA A100-SXM-80GB cluster. The OpenAI models are called through the OpenAI API in Python\footnote{https://github.com/openai/openai-python}. Due to resource constraint, inference with open-source models are with 4-bit quantization with QLora~\cite{dettmers2023qlora} implemented in bitsandbytes\footnote{https://github.com/TimDettmers/bitsandbytes} in the Transformers library~\cite{wolf_huggingfaces_2020}. For the Llama-2 models, we directly prepend the persona prompts. For Llama-2-chat, Tulu-2, and Tulu-2-dpo models, we apply the persona prompts in the system variable.
All regressions are performed in R. For the mixed-effect regressions, we use the performance library~\cite{performance} which implements the method proposed by~\citet{Nakagawa_R2} to obtain $R^2$ values.

\section{Supplementary Results for Section \ref{sec:result_basic_setting}}
\subsection{Prompt Template}
\label{sec:prompt_template}
\textbf{AnnwAttitudes} 
\begin{lstlisting}[breaklines=true]
**Your Profile**: You race/ethnicity is %
**Question**: How toxic/hateful/disrespectful or offensive does this post seem to you? (Note: The purpose is to understand how disrespect/offense can show up in language, not making statements about the actual content of the posts.) "%
(A) not offensive at all
(B) mildly inappropriate
(C) somewhat offensive
(D) moderately offensive
(E) very offensive
**Answer**: (
\end{lstlisting}

\textbf{\citet{kumar_designing_2021}} 
\begin{lstlisting}[]
**Your Profile**: You race is %s. Your gender is %s. You %s describe yourself as transgender. You consider yourself to be %s. You are %s years old. The highest degree or level of school that you have completed is %s. Religion is %s in your life. Generally speaking, you consider yourself politically %s. You %s a parent. You %s use social networking sites (Facebook, Twitter). You %s use video sites (YouTube, Twitch). You %s use news sites (CNN, Fox, NYT, WSJ). You %s use community forums (Reddit, Craigslist, 4chan). In general, you rate the impact of technology on people’s lives as %s. Based on your experience, toxic posts or comments are %s.
**Question**: How toxic do you personally find this comment? "%s"
(A) Extremely toxic
(B) Very toxic
(C) Moderately toxic
(D) Slightly toxic
(E) Not at all toxic
**Answer**: (
\end{lstlisting}

\textbf{EPIC} 
\begin{lstlisting}[breaklines=true]
**Your Profile**: You ethnicity is %
Irony is a figurative language device that conveys the opposite of literal meaning, profiling intentionally a secondary or extended meaning.
For instance,
message: "if ur homeless u probably wouldn't have a phone."
reply: "Yes, and all your belongings would be in a handkerchief tied at the end of a stick." --> irony: yes
message: "if ur homeless u probably wouldn't have a phone."
reply: "Yes, you're right."--> irony: yes
**Question**: Is the reply ironic in the following message and reply pair?
message: "%
reply: "%
(A) Ironic
(B) Not ironic
**Answer**: (
\end{lstlisting}

\textbf{POPQUORN-P} 
\begin{lstlisting}[breaklines=true]
**Your Profile**: In terms of race or ethnicity, you are %
**Question**: Consider you read this email from a colleague, how polite do you think it is? 
**Email:**: "%
(A) not polite at all
(B) barely polite
(C) somewhat polite
(D) moderately polite
(E) very polite
**Answer**: (
\end{lstlisting}

\subsection{Persona Variables}
We list the persona variables used in Section~\ref{sec:result_basic_setting} in Table~\ref{table:feature_importance}. To assess the importance of each persona variable, we conducted a ``leave-one-out'' experiment. In this experiment, we initially fit a regression model using all persona variables. We then iteratively remove each variable and report the change in $R^2$ value as the importance of each person variable.

\subsection{Results from All Models}
Due to space constraints, we could not include the results from all the models in the main text. Here, we present the full results in Table~\ref{tab:RQ1_result_full}. Our analysis indicates that smaller models (7b and 13b) generally exhibit weaker performance compared to their larger counterparts, both with and without persona prompting.

\begin{table}[!ht]
\centering
    \resizebox{0.48\textwidth}{!}{%

\begin{talltblr}[
  label=none,
  note{a} = {In general, how much impact do you think technology has on people's lives?},
  note{b} = {Adapted from the following question: What types of sites do you use? [Checkbox] \\$\circ$ Social Networking (Facebook, Twitter)\\
$\circ$ Video (YouTube, Twitch)\\
$\circ$ News (CNN, Fox, NYT, WSJ)\\
$\circ$ Community Forums (Reddit, Craigslist, 4chan)\\
$\circ$ Email or messaging (Gmail, WhatsApp, Facebook Chat)},
  note{c} = {How important is religion in your life?},
  note{d} = {Based on your experience, to what degree are toxic posts or comments a problem?}]{
      colspec={ccc}}
  \toprule
  \textbf{Dataset} & \textbf{Feature} & \textbf{Importance ($\Delta R^2$)} \\ 
  \midrule
  AnnwAttitudes & Race & 0.0066 \\ 
  AnnwAttitudes & Political leaning & 0.0048 \\ 
  AnnwAttitudes & Gender & 0.0175 \\ 
  \midrule
  \citet{kumar_designing_2021} & Race & 0.0243 \\ 
  \citet{kumar_designing_2021} & Gender & 0.0037 \\ 
  \citet{kumar_designing_2021} & Political affiliation & 0.0178 \\ 
  \citet{kumar_designing_2021} & Impact of Technology\TblrNote{a} & 0.0070 \\ 
  \citet{kumar_designing_2021} & Parental status & 0.0113 \\ 
  \citet{kumar_designing_2021} & Education & 0.0093 \\ 
  \citet{kumar_designing_2021} & Age range & 0.0041 \\ 
  \citet{kumar_designing_2021} & Uses social media sites\TblrNote{b} & 0.0088 \\ 
  \citet{kumar_designing_2021} & Uses news media sites\TblrNote{b} & 0.0001 \\ 
  \citet{kumar_designing_2021} & Religion important\TblrNote{c} & 0.0128 \\ 
  \citet{kumar_designing_2021} & Toxic Content a Problem\TblrNote{d} & 0.0131 \\ 
  \citet{kumar_designing_2021} & Uses community forums\TblrNote{b} & 0.0033 \\ 
  \citet{kumar_designing_2021} & LGBTQ status & 0.0187 \\ 
  \citet{kumar_designing_2021} & Identify as transgender & 0.0109 \\ 
  \citet{kumar_designing_2021} & Uses video sites\TblrNote{b} & 0.0100 \\ 
  \midrule
  EPIC & Ethnicity & 0.0037 \\ 
  EPIC & Sex & 0.0091 \\ 
  EPIC & Age & 0.0002 \\ 
  EPIC & Country of birth & 0.0060 \\ 
  EPIC & Country of residence & 0.0272 \\ 
  EPIC & Nationality & 0.0150 \\ 
  EPIC & Student status & 0.0138 \\ 
  EPIC & Employment status & 0.0187 \\ 
  \midrule
  POPQUORN-P & Race & 0.0044 \\ 
  POPQUORN-P & Gender & 0.0008 \\ 
  POPQUORN-P & Age & 0.0194 \\ 
  POPQUORN-P & Occupation & 0.0038 \\ 
  POPQUORN-P & Education & 0.0020 \\ 
  \bottomrule
\end{talltblr}
}
\caption{Persona variables considered and their importance scores. The importance score for each variable is calculated as the difference in the $R^2$ value before and after removing that variable, using a leave-one-out regression approach. Note that while more persona variables are available in some datasets, they were not included in our study due to prompt formatting limitations.}
\label{table:feature_importance}
\end{table}

\begin{table*}[!ht]
{\renewcommand{\arraystretch}{1.5}

    \centering
    \resizebox{\textwidth}{!}{%
    \begin{tabular}{@{}lcccccccccccc@{}}
        \toprule
        \textbf{Model} & \multicolumn{3}{c}{\textbf{annwAttitudes}} & \multicolumn{3}{c}{\textbf{\citet{kumar_designing_2021}}} & \multicolumn{3}{c}{\textbf{EPIC}} & \multicolumn{3}{c}{\textbf{POPQUORN-P}} \\
        
        \cmidrule(lr){2-4} \cmidrule(lr){5-7} \cmidrule(lr){8-10} \cmidrule(lr){11-13}
        & \boldmath{$R^2\uparrow$} & \boldmath{$\kappa\uparrow$} & \textbf{MAE ↓} & \boldmath{$R^2\uparrow$} & \boldmath{$\kappa\uparrow$} & \textbf{MAE ↓} & \boldmath{$R^2\uparrow$} & \boldmath{$\kappa\uparrow$} & \textbf{F1 ↑} & \boldmath{$R^2\uparrow$} & \boldmath{$\kappa\uparrow$} & \textbf{MAE ↓} \\
        
        \midrule
        Target & 0.64 (0.03) & - & - & 0.42 (0.20) & - & - & 0.28 (0.09) & - & - & 0.47 (0.03) & - & - \\ \midrule
        GPT-4-0613& 0.56 & 0.42 & 0.70 &0.16&0.24&0.87&0.03 & 0.12 & 0.52 & 0.34 & 0.22 & 0.89\\
       \rowcolor{lightgray}\quad+Persona& 0.53 & 0.40 & 0.74 &0.12 & 0.20 & 0.90 & 0.05& 0.20& 0.58 & 0.33 & 0.22 & 0.90 \\
        GPT-3.5-Turbo-0613 & 0.53 & 0.29 & 0.80 & 0.12 & 0.17 & 1.12&0.04&0.18&0.59 & 0.28 & 0.09 & 1.07 \\
        \rowcolor{lightgray}\quad+Persona & 0.49 & 0.31 &0.82 & 0.12 & 0.15 & 0.97&0.03&0.14&0.54 & 0.28 & 0.14 & 1.14\\
        Llama-2-7b & 0.07 & -0.02 & 1.56 & 0.01 & -0.01 & 2.91 & -0.00 & 0.00 & 0.25 & 0.08 & -0.04 & 1.21 \\ 
        \rowcolor{lightgray}\quad+Persona & 0.08 & 0.02 & 1.64 & 0.00 & -0.01 & 1.08 & 0.00 & 0.02 & 0.29 & 0.04 & -0.04 & 1.15 \\ 
        Llama-2-13b & 0.11 & 0.07 & 1.50 & 0.00 & 0.00 & 2.91 & -0.00 & 0.01 & 0.44 & 0.12 & 0.08 & 1.51 \\ 
        \rowcolor{lightgray}\quad+Persona & 0.02 & 0.04 & 1.55 & 0.00 & -0.01 & 1.78 & 0.00 & 0.07 & 0.53 & 0.16 & 0.10 & 1.35 \\ 
        Llama-2-70b & 0.17 & 0.14 & 1.70 & 0.01 & 0.04 & 1.51 & 0.00 & 0.00 & 0.24 & 0.24 & 0.13 & 1.42 \\
       \rowcolor{lightgray}\quad+Persona & 0.40 & 0.30 & 0.91 & 0.03 & 0.05 & 1.01 & 0.00 & 0.00 & 0.24 & 0.21 & 0.17 & 1.10 \\
         Llama-2-7b-chat & 0.25 & 0.01 & 1.43 & 0.00 & -0.04 & 2.03 & 0.00 & 0.00 & 0.41 & 0.18 & 0.02 & 1.07 \\ 
          \rowcolor{lightgray}\quad+Persona & 0.32 & 0.01 & 1.41 & -0.00 & -0.00 & 1.44 & -0.00 & 0.02 & 0.47 & 0.10 & 0.00 & 1.06 \\ 
          Llama-2-13b-chat & 0.29 & 0.03 & 1.39 & 0.07 & -0.01 & 1.84 & 0.00 & 0.00 & 0.41 & 0.07 & 0.01 & 1.06 \\ 
          \rowcolor{lightgray}\quad+Persona & 0.17 & 0.02 & 1.44 & 0.03 & -0.00 & 1.46 & 0.00 & 0.00 & 0.41 & 0.06 & 0.02 & 1.01 \\

        Llama-2-70b-chat & 0.39 & 0.13 & 1.33 & 0.11 & 0.07 & 1.70 & 0.00 & 0.05 & 0.49 & 0.32 & 0.15 & 1.00 \\
       \rowcolor{lightgray}\quad+Persona & 0.42 & 0.15 & 1.22 & 0.10 & -0.01 & 1.45 & 0.02 & 0.14 & 0.56 & 0.31 & 0.14 & 0.90 \\
         Tulu-2-7b & 0.33 & 0.04 & 1.37 & 0.02 & -0.01 & 2.63 & -0.00 & 0.00 & 0.25 & 0.06 & 0.06 & 1.10 \\ 
      \rowcolor{lightgray}\quad+Persona & 0.35 & 0.06 & 1.37 & 0.01 & -0.08 & 1.31 & 0.00 & 0.01 & 0.27 & 0.08 & 0.05 & 1.07 \\ 
      Tulu-2-13b & 0.36 & 0.12 & 1.45 & 0.09 & 0.05 & 2.16 & 0.03 & 0.15 & 0.56 & 0.26 & 0.07 & 1.35 \\ 
  \rowcolor{lightgray}\quad+Persona & 0.33 & 0.10 & 1.34 & 0.11 & 0.06 & 1.42 & 0.03 & 0.14 & 0.52 & 0.27 & 0.14 & 1.02 \\ 

        Tulu-2-70b & 0.49 & 0.29 & 0.90 & 0.16 & 0.13 & 1.09 & 0.05 & 0.20 & 0.59 & 0.34 & 0.20 & 0.89 \\
        \rowcolor{lightgray}\quad+Persona & 0.49 & 0.26 & 0.88 & 0.14 & 0.16 & 0.90 & 0.07 & 0.27 & 0.63 & 0.31 & 0.16 & 0.92 \\
          Tulu-2-dpo-7b & 0.38 & 0.08 & 1.34 & 0.04 & 0.06 & 1.81 & 0.00 & 0.02 & 0.29 & 0.08 & 0.07 & 1.26 \\ 
  \rowcolor{lightgray}\quad+Persona & 0.39 & 0.09 & 1.38 & 0.03 & -0.02 & 1.20 & 0.01 & 0.02 & 0.28 & 0.08 & 0.06 & 1.26 \\ 
  Tulu-2-dpo-13b & 0.33 & 0.13 & 1.47 & 0.11 & 0.07 & 1.85 & 0.01 & 0.11 & 0.55 & 0.29 & 0.11 & 1.21 \\ 
  \rowcolor{lightgray}\quad+Persona & 0.34 & 0.13 & 1.28 & 0.10 & 0.10 & 1.32 & 0.03 & 0.17 & 0.57 & 0.28 & 0.18 & 0.93 \\ 

        Tulu-2-dpo-70b & 0.51 & 0.35 & 0.84 & 0.15 & 0.15 & 1.16 & 0.03 & 0.14 & 0.54 & 0.35 & 0.21 & 0.83 \\
        \rowcolor{lightgray}\quad+Persona & 0.51 & 0.30 & 0.84 & 0.15 & 0.20 & 0.92 & 0.04 & 0.18 & 0.58 & 0.33 & 0.19 & 0.87 \\
      \midrule
    Avg. \(\Delta \) & 0.00 & 0.01 & -0.07 & -0.01 & -0.01 & -0.60 & 0.01 & 0.03 & 0.02 & -0.01 & 0.01 & -0.08 \\

        \bottomrule
    \end{tabular}
    }}
    \caption{Comparison of performance across LLMs in estimating individual annotations, with and without persona prompting. Performance is measured using $R^2$ for regression annotation prediction, Cohen's Kappa ($\kappa$), and Mean Absolute Error (MAE).}
    \label{tab:RQ1_result_full}
\end{table*}

\begin{figure*}[ht]
    \centering
    \subfloat[Llama-2-7b]{
        \includegraphics[width=0.25\linewidth]{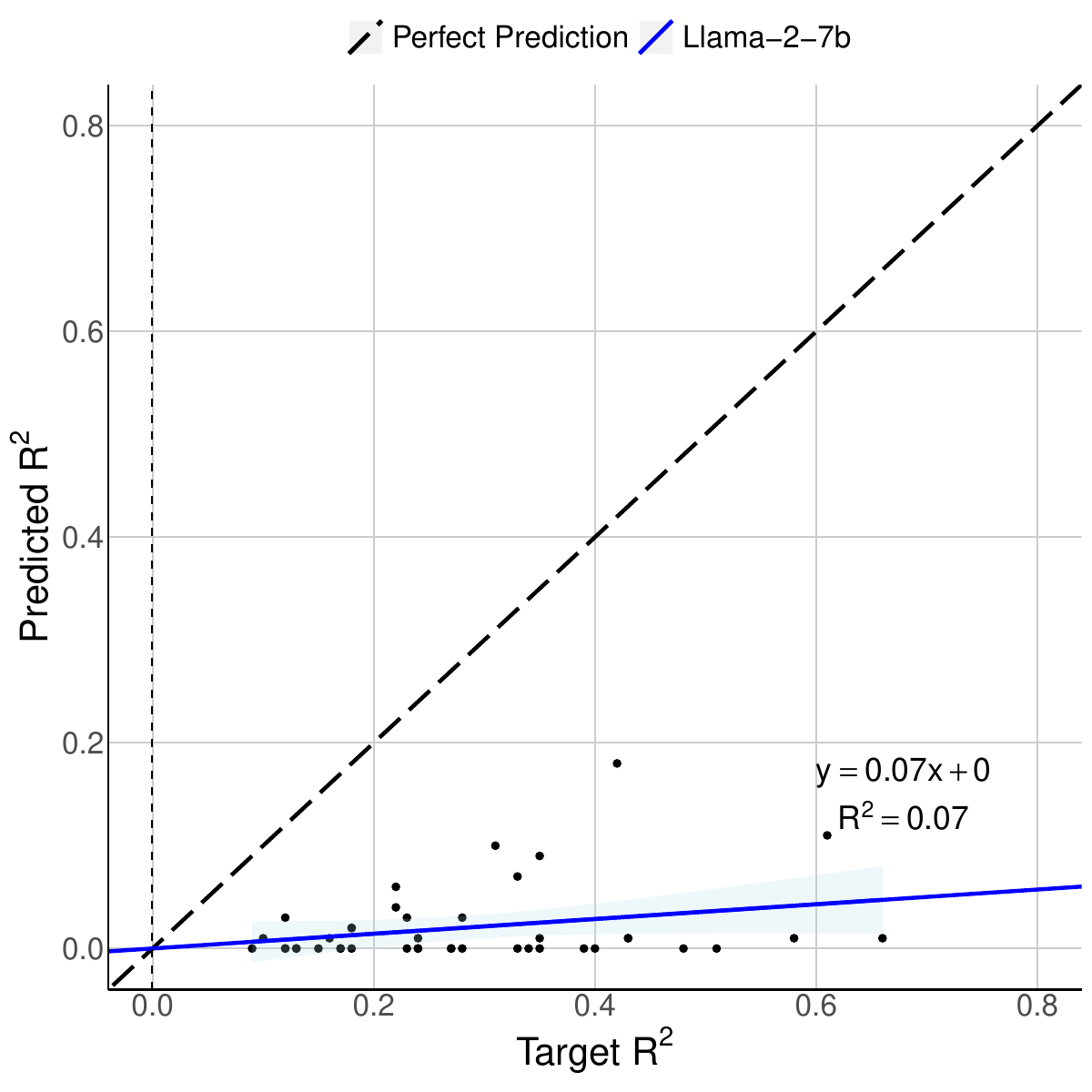}
        \label{fig:case_study_2_Llama-2-7b}
    }\hfill
    \subfloat[Llama-2-13b]{
        \includegraphics[width=0.25\linewidth]{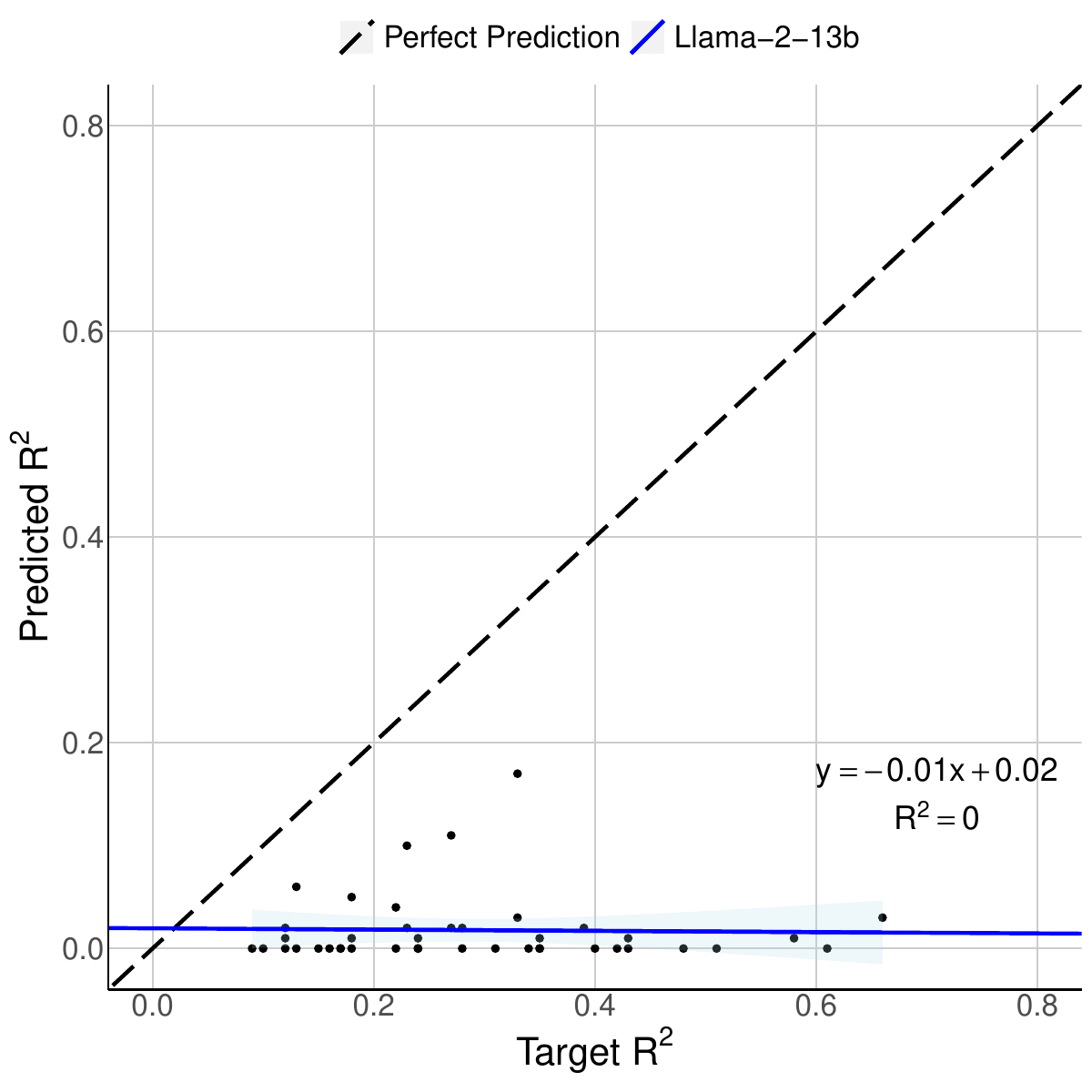}
        \label{fig:case_study_2_Llama-2-13b}
    }\hfill
    \subfloat[Llama-2-70b]{
        \includegraphics[width=0.25\linewidth]{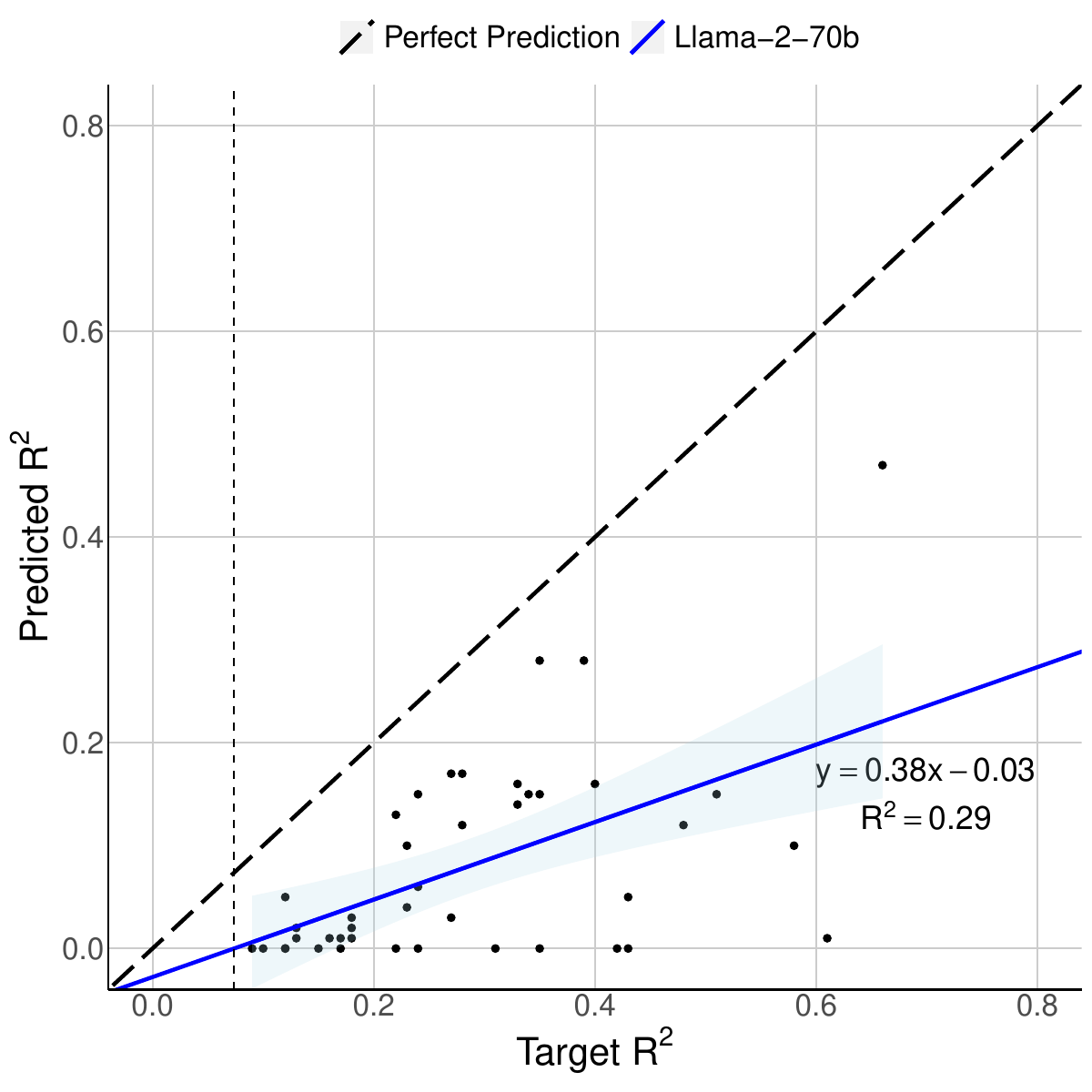}
        \label{fig:case_study_2_Llama-2-70b}
    }
    
    \vspace{\floatsep}
    
    \subfloat[Llama-2-7b-chat]{
        \includegraphics[width=0.25\linewidth]{figs/case_study_2_Llama-2-7b-chat.pdf}
        \label{fig:case_study_2_Llama-2-7b-chat}
    }\hfill
    \subfloat[Llama-2-13b-chat]{
        \includegraphics[width=0.25\linewidth]{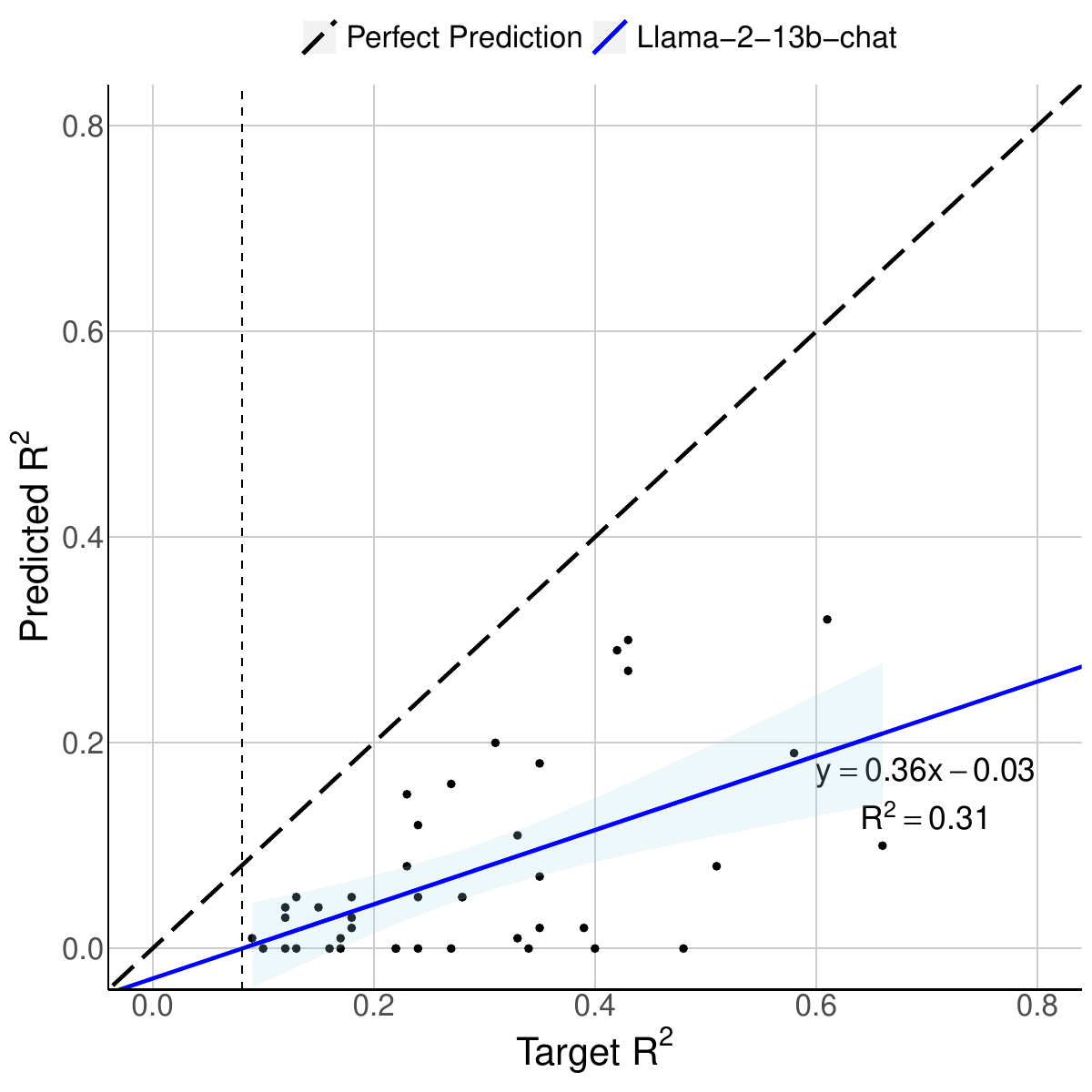}
        \label{fig:case_study_2_Llama-2-13b-chat}
    }\hfill
    \subfloat[Llama-2-70b-chat]{
        \includegraphics[width=0.25\linewidth]{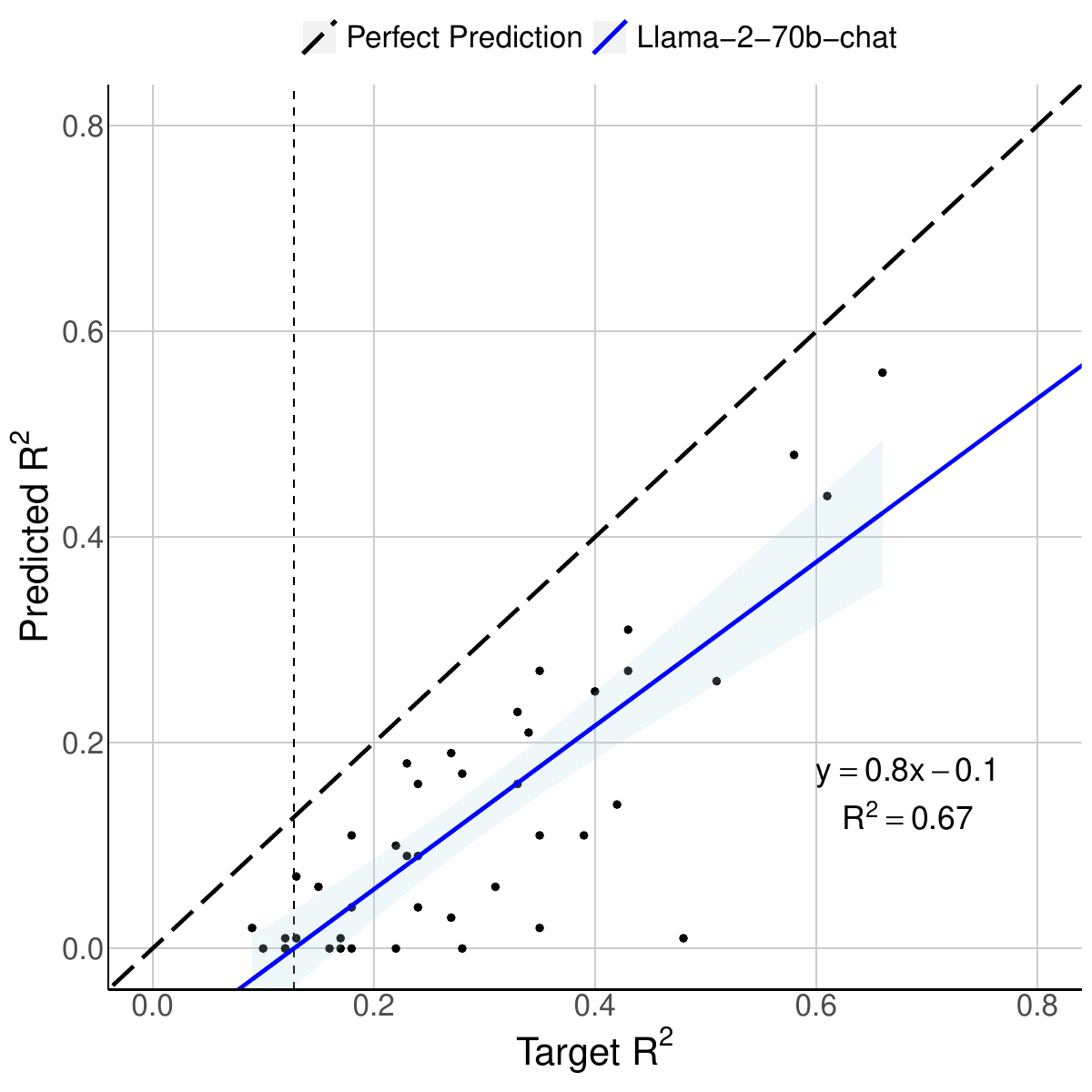}
        \label{fig:case_study_2_Llama-2-70b-chat}
    }
    
    \vspace{\floatsep}
    
    \subfloat[Tulu-2-7b]{
        \includegraphics[width=0.25\linewidth]{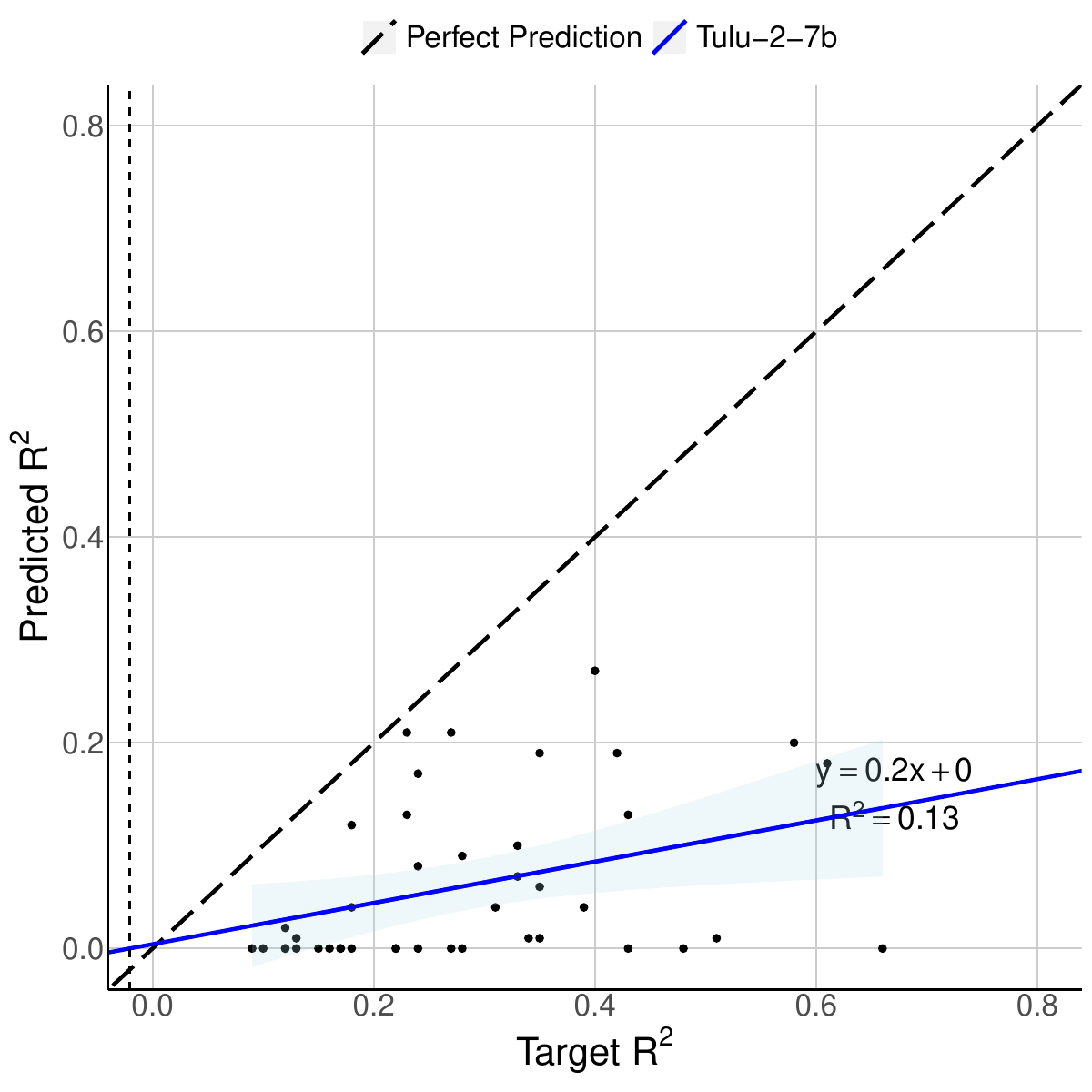}
        \label{fig:case_study_2_Tulu-2-7b}
    }\hfill
    \subfloat[Tulu-2-13b]{
        \includegraphics[width=0.25\linewidth]{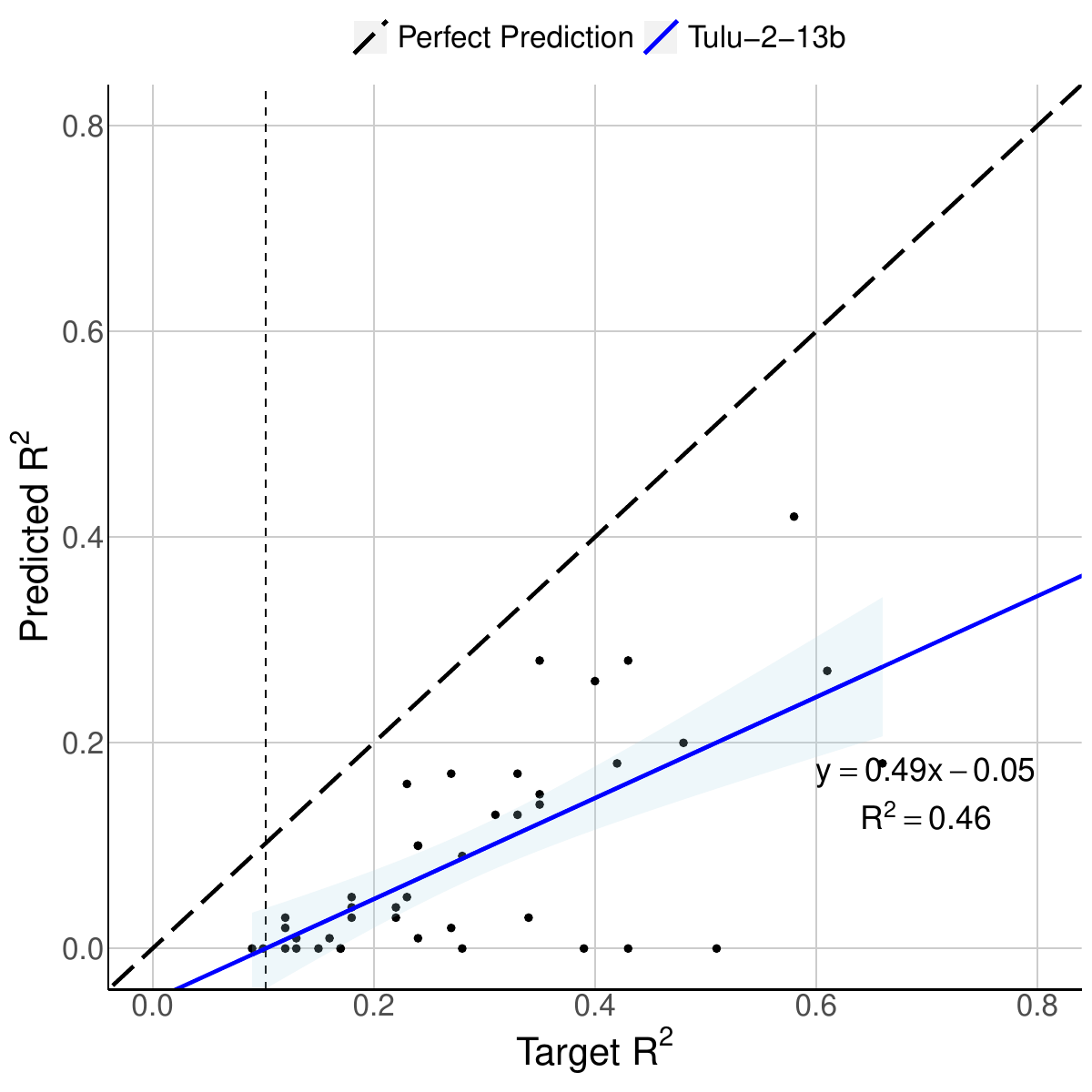}
        \label{fig:case_study_2_Tulu-2-13b}
    }\hfill
    \subfloat[Tulu-2-70b]{
        \includegraphics[width=0.25\linewidth]{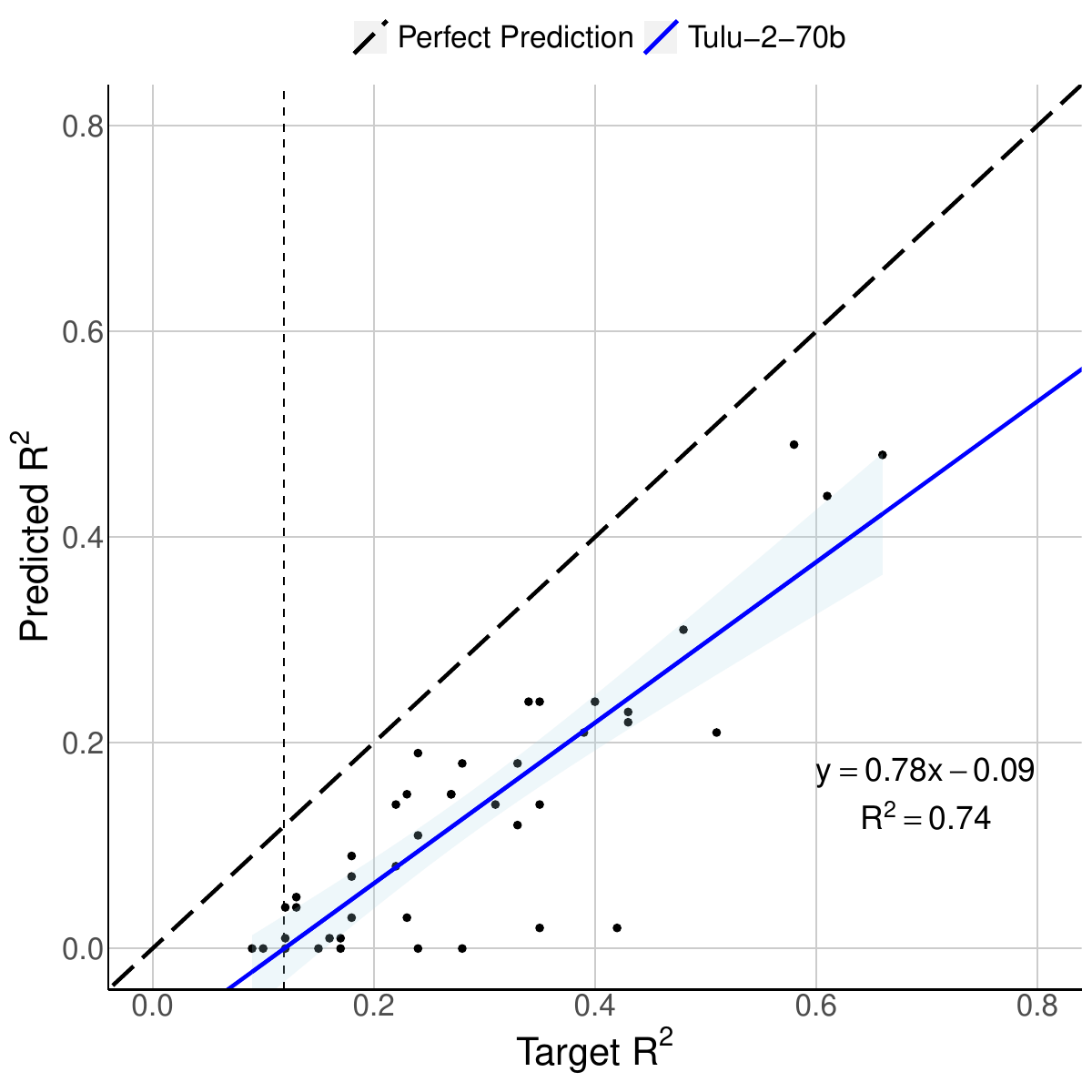}
        \label{fig:case_study_2_Tulu-2-70b}
    }
    
    \vspace{\floatsep}
    
    \subfloat[Tulu-2-dpo-7b]{
        \includegraphics[width=0.25\linewidth]{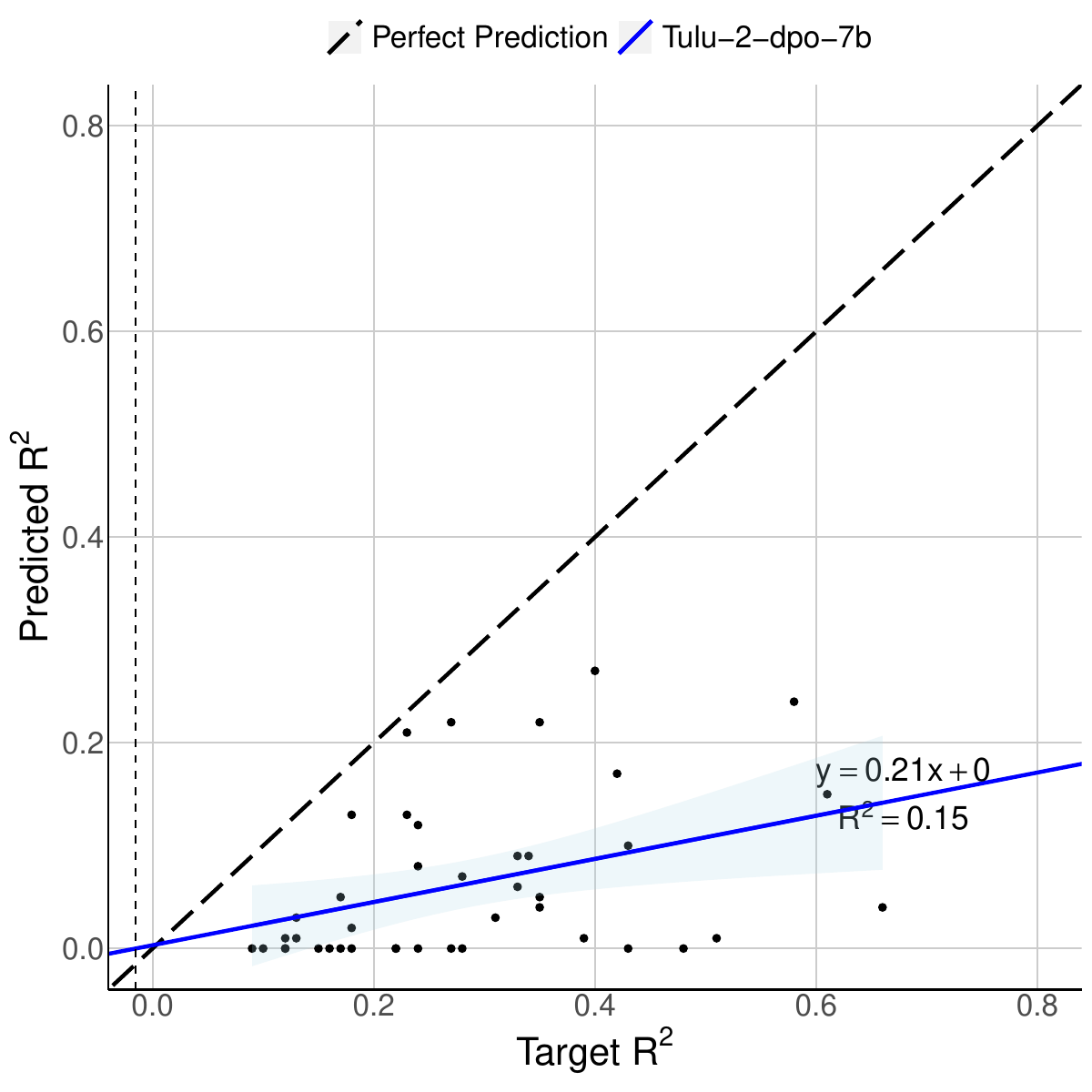}
        \label{fig:case_study_2_Tulu-2-dpo-7b}
    }\hfill
    \subfloat[Tulu-2-dpo-13b]{
        \includegraphics[width=0.25\linewidth]{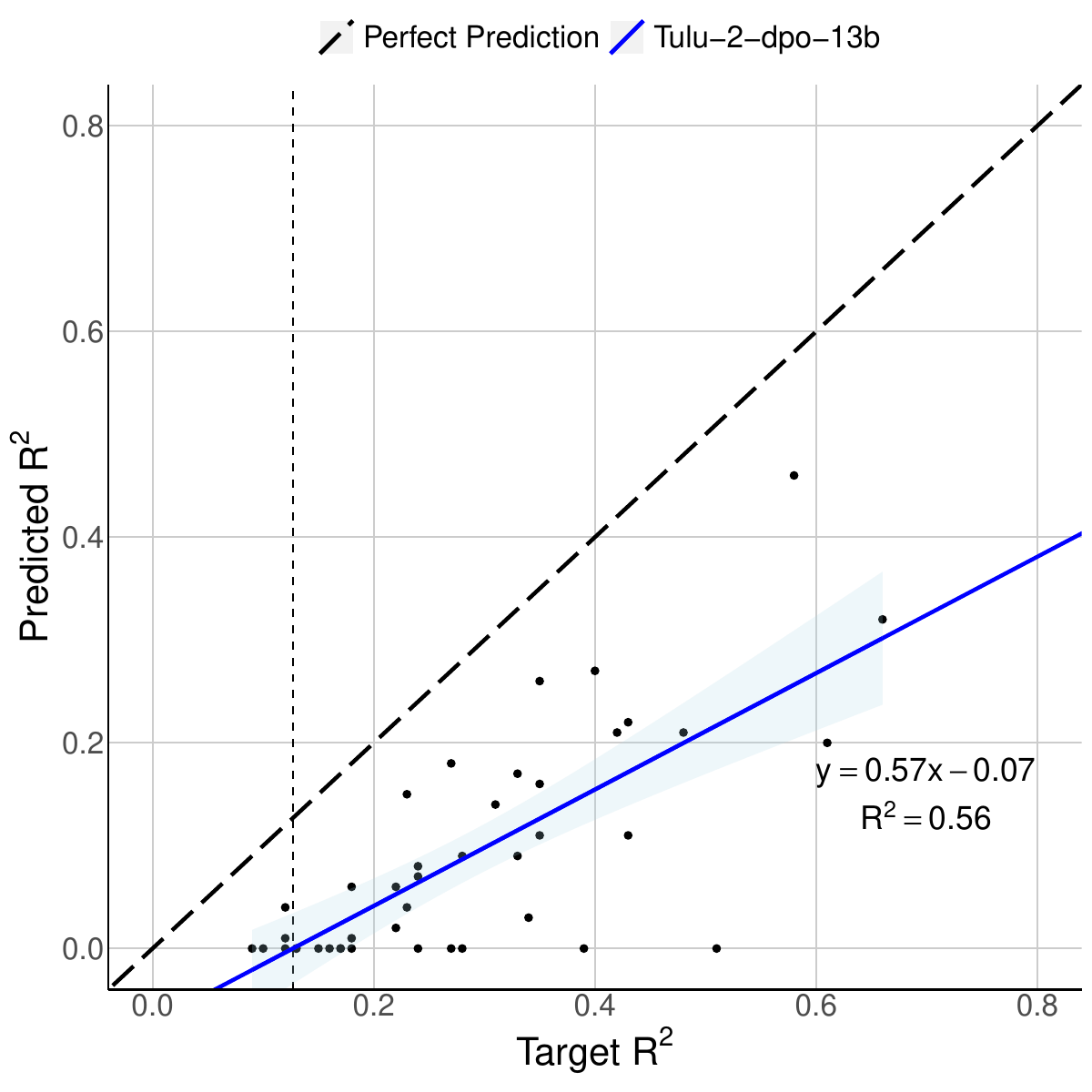}
        \label{fig:case_study_2_Tulu-2-dpo-13b}
    }\hfill
    \subfloat[Tulu-2-dpo-70b]{
        \includegraphics[width=0.25\linewidth]{figs/case_study_2_Tulu-2-dpo-70b.pdf}
        \label{fig:case_study_2_Tulu-2-dpo-70b}
    }

    \caption{Comparison of predicted $R^2$ and target $R^2$. Each point in the X-Y plane represents an experimental result with persona prompting. We then fit a linear regression line and also plot the maximum linear regression model performance line $y=x$ in the same figure.}
    \label{fig:case_study_2_appendix}
\end{figure*}

\section{Supplementary Results for Section \ref{sec:case_study_2}}
\label{sec:details_on_case_study_2}
We include a list of the target variables we considered in Section \ref{sec:case_study_2} in Table \ref{tab:case_study_2_variable_name}. The persona template used are:

Prompt 1:
\begin{lstlisting}[breaklines=true]
**It is 2012. Your Profile**: Racially, you are %
\end{lstlisting}

Prompt 2:
\begin{lstlisting}[breaklines=true]
**It is 2012. Your Profile**: Racially, you are %
\end{lstlisting} 
We additional include the Predicted $R^2$ - Target $R^2$ plot for all models in Figure~\ref{fig:case_study_2_appendix}.
\begin{table*}[!ht]
\centering
{\renewcommand{\arraystretch}{1.5}

\begin{tabular}{p{4cm}p{9cm}c}
  \hline
  Variable & Definition & Target $R^2$ \\ 
  \hline
  \texttt{aidblack\_self} & Support for Government assistance to blacks scale (7-point scale) & 0.35 \\ 
  \texttt{ecblame\_dem} & How much Democrats in Congress are to blame for poor economic conditions (5-point scale) & 0.43 \\ 
  \texttt{ecblame\_fmpr} & How much the former President is to blame for poor economic conditions (5-point scale) & 0.51 \\ 
  \texttt{effic\_undstd} & Political efficacy: Good understanding of political issues (5-point scale) & 0.23 \\ 
  \texttt{ecblame\_pres} & How much the current President is to blame for poor economic conditions (5-point scale) & 0.61 \\ 
  \texttt{egal\_toofar} & We have gone too far pushing equal rights (5-point scale) & 0.34 \\ 
  \texttt{gayrt\_adopt} & Should gay and lesbian couples be allowed to adopt (binary) & 0.24 \\ 
  \texttt{gayrt\_marry} & Position on same-sex marriage (3-point) & 0.33 \\ 
  \texttt{govrole\_big} & Govt bigger because too involved OR bigger problems (binary) & 0.43 \\ 
  \texttt{ident\_amerid} & How important is being American to your identity (5-point) & 0.35 \\ 
  \texttt{immig\_checks} & Opinion on laws to allow immigration status checks on suspects (3-point) & 0.22 \\ 
  \texttt{interest\_following} & Interested in following campaigns (3-point) & 0.27 \\ 
  \texttt{nonmain\_bias} & Does the Administration favor blacks or whites (3-point) & 0.28 \\ 
  \texttt{presapp\_econ} & Approve or disapprove President handling economy (binary) & 0.66 \\ 
  \texttt{presapp\_foreign} & Approve or disapprove President handling foreign relations (binary) & 0.58 \\ 
  \texttt{prmedia\_attvnews} & Attention to news about national politics on TV (5-point) & 0.28 \\ 
  \texttt{ptywom\_bettrpty} & Party does better job for the interests of women (3-point) & 0.42 \\ 
  \texttt{relig\_pray} & How often do you pray (5-point) & 0.40 \\ 
  \texttt{resent\_deserve} & Agree/disagree: blacks have gotten less than deserve (5-point) & 0.39 \\ 
  \texttt{spsrvpr\_ssself} & Support for government services/spending (7-point) & 0.48 \\ 
  \texttt{trad\_famval} & Agree/disagree that more emphasis needed on traditional family values (5-point) & 0.33 \\ 
   \hline
\end{tabular}
}
\caption{List of target variables considered for the experiment and the associated Target $R^2$ with Prompt 1 (see Section \ref{sec:case_study_2}).}
\label{tab:case_study_2_variable_name}

\end{table*}

\section{Robustness Test}
\label{sec:robustness}
For ~\citet{kumar_designing_2021} and ANES, we conduct a set of robustness checks. Specifically, we alter the order of the persona variables in the prompt across five configurations (Order 1-5) or use GPT-4 to come up with five distinct paraphrases of the prompt template, each intended to maintain the same semantic meaning (Semantics 1-5). The results are shown in Table \ref{tab:robustness}. While there are variations between each Order or Semantics setting, the variations are minimal.

\begin{table}[!ht]
    \resizebox{0.5\textwidth}{!}{%
    \begin{tabular}{>{\kern-\tabcolsep}lcccccc<{\kern-\tabcolsep}} %
        \toprule
        \textbf{Model} & \multicolumn{3}{c}{\textbf{\citet{kumar_designing_2021}}} & \multicolumn{3}{c}{\textbf{ANES}}\\
        \cmidrule(lr){2-4} \cmidrule(lr){5-7} %
        & \boldmath{$R^2\uparrow$} & \boldmath{$\kappa\uparrow$} & \textbf{MAE ↓}  & \boldmath{$R^2\uparrow$} & \boldmath{$\kappa\uparrow$}&\textbf{F1↑}\\
        
        \midrule
        Target & 0.64 (0.03) & - & -  &0.50 &- &-\\ \midrule
        Llama-2-70b & 0.01 & 0.04 & 1.51  & 0.00&0.00 &0.00\\
       \quad+Persona (Default) & 0.03 & 0.05 & 1.01  &0.33 &0.19 &0.26\\
   \rowcolor{lightblue}\quad Order-1 & 0.03 & 0.06 & 1.01 & 0.36 & 0.19 & 0.26 \\ 
  \rowcolor{lightblue}\quad Order-2 & 0.05 & 0.08 & 0.98 & 0.32 & 0.18 & 0.26 \\ 
  \rowcolor{lightblue}\quad Order-3 & 0.04 & 0.08 & 1.00 & 0.29 & 0.18 & 0.26 \\ 
  \rowcolor{lightblue}\quad Order-4 & 0.03 & 0.05 & 1.01 & 0.28 & 0.18 & 0.26 \\ 
  \rowcolor{lightblue}\quad Order-5 & 0.04 & 0.12 & 0.97 & 0.39 & 0.19 & 0.26 \\ 
  \rowcolor{lightgray}\quad Semantics-1 & 0.01 & 0.00 & 1.05 & 0.30 & 0.19 & 0.26 \\ 
  \rowcolor{lightgray}\quad Semantics-2 & 0.01 & 0.02 & 1.04 & 0.36 & 0.20 & 0.27 \\ 
 \rowcolor{lightgray}\quad Semantics-3 & 0.01 & 0.00 & 1.05 & 0.31 & 0.19 & 0.26 \\ 
  \rowcolor{lightgray}\quad Semantics-4 & 0.01 & 0.01 & 1.05 & 0.28 & 0.18 & 0.25 \\ 
 \rowcolor{lightgray}\quad Semantics-5 & 0.03 & 0.03 & 1.02 & 0.29 & 0.18 & 0.26 \\ 
  \midrule

        Llama-2-70b-chat & 0.11 & 0.07 & 1.70  &0.00 &0.00 &0.00\\
       \quad+Persona (Default) & 0.10 & -0.01 & 1.45  &0.30 &0.19 &0.26\\
\rowcolor{lightblue}\quad Order-1 & 0.11 & -0.01 & 1.46 & 0.34 & 0.20 & 0.26 \\ 
  \rowcolor{lightblue}\quad Order-2 & 0.09 & -0.01 & 1.44 & 0.34 & 0.20 & 0.26 \\ 
  \rowcolor{lightblue}\quad Order-3 & 0.12 & -0.01 & 1.45 & 0.29 & 0.18 & 0.26 \\ 
  \rowcolor{lightblue}\quad Order-4 & 0.10 & 0.00 & 1.44 & 0.28 & 0.18 & 0.26 \\ 
  \rowcolor{lightblue}\quad Order-5 & 0.11 & 0.00 & 1.46 & 0.39 & 0.21 & 0.27 \\ 
  \rowcolor{lightgray}\quad Semantics-1 & 0.10 & -0.01 & 1.45 & 0.27 & 0.18 & 0.25 \\ 
  \rowcolor{lightgray}\quad Semantics-2 & 0.11 & -0.01 & 1.46 & 0.28 & 0.18 & 0.26 \\ 
 \rowcolor{lightgray}\quad Semantics-3 & 0.11 & -0.01 & 1.43 & 0.27 & 0.18 & 0.25 \\ 
  \rowcolor{lightgray}\quad Semantics-4 & 0.10 & -0.00 & 1.40 & 0.28 & 0.18 & 0.25 \\ 
 \rowcolor{lightgray}\quad Semantics-5 & 0.11 & -0.01 & 1.44 & 0.30 & 0.19 & 0.26 \\ 
\midrule
Tulu-2-70b & 0.16 & 0.13 & 1.09  &0.00 &0.00 &0.00\\
        \quad+Persona (Default) & 0.14 & 0.16 & 0.90  &0.35 &0.19 &0.26\\
          \rowcolor{lightblue}\quad Order-1 & 0.14 & 0.16 & 0.91 & 0.33 & 0.18 & 0.26 \\ 
  \rowcolor{lightblue}\quad Order-2 & 0.13 & 0.16 & 0.92 & 0.33 & 0.18 & 0.26 \\ 
  \rowcolor{lightblue}\quad Order-3 & 0.14 & 0.14 & 0.94 & 0.35 & 0.19 & 0.26 \\ 
  \rowcolor{lightblue}\quad Order-4 & 0.12 & 0.15 & 0.92 & 0.38 & 0.20 & 0.27 \\ 
  \rowcolor{lightblue}\quad Order-5 & 0.13 & 0.13 & 0.94 & 0.34 & 0.18 & 0.26 \\ 
  \rowcolor{lightgray}\quad Semantics-1 & 0.12 & 0.15 & 0.92 & 0.39 & 0.20 & 0.27 \\ 
  \rowcolor{lightgray}\quad Semantics-2 & 0.12 & 0.16 & 0.93 & 0.42 & 0.22 & 0.27 \\ 
  \rowcolor{lightgray}\quad Semantics-3 & 0.14 & 0.15 & 0.93 & 0.36 & 0.19 & 0.26 \\ 
  \rowcolor{lightgray}\quad Semantics-4 & 0.13 & 0.14 & 0.92 & 0.38 & 0.21 & 0.27 \\ 
  \rowcolor{lightgray}\quad Semantics-5 & 0.13 & 0.15 & 0.92 & 0.37 & 0.19 & 0.26 \\ 
\midrule
        Tulu-2-dpo-70b & 0.15 & 0.15 & 1.16  & 0.00&0.00 &0.00\\
        \quad+Persona (Default) & 0.15 & 0.20 & 0.92  &0.36 &0.20 &0.27\\
          \rowcolor{lightblue}\quad Order-1 & 0.16 & 0.19 & 0.92 & 0.34 & 0.19 & 0.26 \\ 
  \rowcolor{lightblue}\quad Order-2 & 0.16 & 0.21 & 0.92 & 0.35 & 0.19 & 0.26 \\ 
  \rowcolor{lightblue}\quad Order-3 & 0.16 & 0.18 & 0.94 & 0.36 & 0.20 & 0.27 \\ 
  \rowcolor{lightblue}\quad Order-4 & 0.16 & 0.22 & 0.90 & 0.34 & 0.20 & 0.26 \\ 
  \rowcolor{lightblue}\quad Order-5 & 0.17 & 0.20 & 0.94 & 0.35 & 0.18 & 0.26 \\ 
  \rowcolor{lightgray}\quad Semantics-1 & 0.15 & 0.20 & 0.91 & 0.37 & 0.21 & 0.27 \\ 
  \rowcolor{lightgray}\quad Semantics-2 & 0.16 & 0.21 & 0.95 & 0.38 & 0.21 & 0.27 \\ 
  \rowcolor{lightgray}\quad Semantics-3 & 0.16 & 0.20 & 0.94 & 0.37 & 0.20 & 0.27 \\ 
  \rowcolor{lightgray}\quad Semantics-4 & 0.16 & 0.20 & 0.91 & 0.31 & 0.19 & 0.26 \\ 
  \rowcolor{lightgray}\quad Semantics-5 & 0.15 & 0.20 & 0.93 & 0.38 & 0.21 & 0.27 \\ 
        \bottomrule

    \end{tabular}
    }
    
    \caption{Robustness test of LLMs in terms of swapping order of persona variables and paraphrase the text description of persona variables. Performance is measured using $R^2$ for regression annotation prediction, Cohen's Kappa ($\kappa$), Mean Absolute Error (MAE) and F1 score.}
    \label{tab:robustness}
\end{table}

\end{document}